\documentclass[journal]{IEEEtran}

\usepackage{amsmath}
\usepackage{amsfonts}
\usepackage{amssymb}
\usepackage{mathtools}

\usepackage{algorithm}
\usepackage{algorithmic}

\usepackage{booktabs}
\usepackage{array}
\usepackage{multirow}
\usepackage[table]{xcolor}
\usepackage{graphicx}
\usepackage[caption=false,font=footnotesize]{subfig}
\usepackage{textcomp}
\usepackage{url}
\usepackage{cite}
\usepackage{microtype}

\usepackage[hidelinks]{hyperref}

\newcommand{\HeatACO}{\textsc{HeatACO}}
\newcommand{\MMAS}{\textsc{MMAS}}
\newcommand{\TSP}{\textsc{TSP}}
\newcommand{\ATSP}{\textsc{ATSP}}

\newcommand{\SigBetter}{\,{\ensuremath{(\uparrow)}}}
\newcommand{\SigTie}{\,{\ensuremath{(=)}}}
\newcommand{\SigWorse}{\,{\ensuremath{(\downarrow)}}}
\newcommand{\TableFont}{\footnotesize\renewcommand{\arraystretch}{1.2}}
\newcommand{\TableNoteFont}{\footnotesize}

\newenvironment{IEEEimpact}
{\begingroup\renewcommand{\abstractname}{Impact Statement}\begin{abstract}}
{\end{abstract}\endgroup}

\definecolor{tabRed}{HTML}{FCBBA1}
\definecolor{tabOrange}{HTML}{FDD0A2}
\definecolor{tabYellow}{HTML}{FFF2CC}
\definecolor{tabGreen}{HTML}{C7E9C0}

\begin{document}

\title{\HeatACO: A Heatmap-Guided Max--Min Ant System for Large-Scale Travelling Salesman Problems}

\author{Bo-Cheng Lin, Yi Mei, Mengjie Zhang%
\thanks{Authors are with the Centre for Data Science and Artificial Intelligence \& the School of Engineering and Computer Science, Victoria University of Wellington, Wellington, 6140, New Zealand.}%
\thanks{Corresponding authors: Bo-Cheng Lin (e-mail: bocheng.lin@ecs.vuw.ac.nz) and Yi Mei (e-mail: yi.mei@ecs.vuw.ac.nz).}}

\markboth{IEEE Transactions ,~Vol.~XX, No.~X, Month~YYYY}%
{HeatACO TSP Decoding}

\maketitle

\begin{abstract}
Non-autoregressive neural solvers predict an edge-confidence heatmap for the Travelling Salesman Problem (TSP) in one forward pass, but a decoder must still produce a feasible Hamiltonian cycle. As instance size grows, this stage must reconcile a quadratic number of edge scores with global tour constraints. Greedy edge merging is fast and deterministic but produces low-quality tours, whereas Monte Carlo Tree Search (MCTS) over ${k}$-opt moves recovers better tours at high computational cost and requires predictor-specific tuning. We propose \HeatACO{}, a predictor-agnostic heatmap-to-tour decoder. Its key is a capped, degree-aware evidence factor that integrates a fixed heatmap into a Max--Min Ant System (\MMAS). The factor rewards only edge confidence beyond a node's tour-degree capacity, and its strength is scaled automatically from the pheromone dynamic range, allowing one configuration to decode heatmaps from different predictors without retraining or per-predictor tuning. Across four heatmap sources, \HeatACO{} produces higher-quality solutions in less decoding time than the MCTS baseline on TSP500, TSP1K and TSP10K. Against matched standard MMAS baselines with the same search budget, heatmap guidance improves construction for all four predictors at both scales and remains beneficial with local search. \HeatACO{} also transfers competitively to several distribution shifts and the asymmetric TSP (ATSP). Our post-hoc analysis identifies measurable heatmap properties associated with the observed performance variation.

\end{abstract}

\begin{IEEEimpact}
Many learning-based methods for large routing problems score promising edges rather than return a complete route. A decoder must then enforce global route constraints, yet existing procedures can be slow or require model- and scale-specific tuning. \HeatACO{} provides a common, automatically scaled procedure for converting heatmaps from different pretrained models into feasible tours without retraining. Across in-distribution TSP benchmarks from 500 to 10,000 nodes, it obtained lower gaps and shorter reported decoding times than the published MCTS baselines in every available same-heatmap comparison. Matched classical-search comparisons show that guidance is most valuable when it contributes structure not recovered by local repair. Directed and distribution-shift experiments identify when learned evidence transfers and when conventional geometric search is preferable, making existing predictors easier to compare and reuse.
\end{IEEEimpact}


\section{Introduction}
\label{sec:introduction}

\IEEEPARstart{T}{he} Travelling Salesman Problem (TSP) is a classic problem in combinatorial optimisation and a standard testbed for new solution methods~\cite{applegate_tsp_computational_study_2006}. It is NP-hard~\cite{korte2011combinatorial}, and the size of its search space grows factorially with the number of nodes, so large instances remain difficult despite decades of progress on exact solvers and local search~\cite{applegate_tsp_computational_study_2006,helsgaun_lkh_2000}. Exact methods do not scale to the largest instances within practical limits \cite{ilavarasi_variants_2014}, and the burden of solving them falls on heuristics, where carefully engineered local search, such as Lin--Kernighan and its extensions, still defines the state of the art at scale~\cite{lin_kernighan_1973,helsgaun_lkh_2000}. More recently, machine learning has emerged as a complementary route to such NP-hard combinatorial problems, learning solution strategies directly from data with the aim of producing high-quality solutions quickly and of transferring across instances~\cite{bengio_machine_2021,karimi_mamaghan_ml_metaheuristics_2022}.

Among these machine-learning approaches, heatmap-based non-autoregressive (NAR) neural solvers~\cite{joshi_efficient_2019,fu_generalize_2021} separate inference into two stages. A predictor computes an edge-confidence heatmap in a single parallel forward pass, whereas autoregressive models construct a tour one node at a time~\cite{kool_attention_2019,baSurveyNeuralRouting2026}, and a decoder then converts the heatmap into a feasible tour. The heatmap assigns every edge a score reflecting how likely that edge is to appear in a high-quality tour, and the predictor is trained so that edges on good tours receive higher scores than the rest~\cite{sun_difusco_2023,min_unsupervised_2023,qiu_dimes_2022}.

A heatmap, however, is not itself a tour~\cite{joshi_efficient_2019}. At large scale, the decoder must select the $n$ edges of a globally consistent tour from $O(n^2)$ heatmap scores. A heatmap typically assigns high confidence to more than two edges at a node, admits disconnected cycles, and contains locally plausible edges that cannot coexist in a single Hamiltonian cycle. Even when the predictor architecture models graph context, its heatmap output and training loss do not by themselves enforce the degree and subtour constraints of a tour~\cite{joshi_efficient_2019,sun_difusco_2023,min_unsupervised_2023}. A node can therefore receive high confidence on many incident edges, none of which the predictor is forced to disambiguate. Its numerical scale is also predictor-dependent, so the same confidence value need not carry the same weight across models, and a given model can produce denser and less separated heatmaps under scale or distribution shift. The values are therefore best read as ordinal evidence rather than as calibrated probabilities. Turning the heatmap into a feasible tour is therefore a distinct second stage within the solver, the heatmap-to-tour decoding problem, and the decoder largely determines whether the learned edge evidence becomes a globally consistent tour~\cite{xia_position_2024,pan_beyond_2025}. The decoder must enforce the degree and subtour constraints, exploit the reliable edges while rejecting high-confidence false positives, and do so fast enough to preserve the parallelism that NAR prediction provides.

Two commonly used fixed-heatmap decoders in this line of work are greedy merging and MCTS-guided $k$-opt search. Greedy merging ranks candidate edges by a combined heatmap and distance score and inserts each edge that keeps the partial solution feasible~\cite{sun_difusco_2023,li_t2t_2023}. It is fast and deterministic, but it builds a single tour in one pass and in practice reaches only low quality, leaving most of the useful information in the heatmap unexploited~\cite{sun_difusco_2023}. MCTS-guided $k$-opt search uses the heatmap to guide repeated local moves and recovers much higher-quality tours~\cite{fu_generalize_2021,browne_survey_2012}, but it requires carefully designed action spaces and predictor-specific tuning of its exploration and rollout parameters~\cite{pan_beyond_2025}. 

Ant colony optimisation (ACO) \cite{dorigo_maniezzo_colorni_ant_system_1996} occupies exactly this middle ground. It is a population-based metaheuristic in which simple agents repeatedly sample feasible tours from a pheromone distribution over edges, after which short tours reinforce the edges they use and the rest evaporate, so that the search refines its own bias over many low-cost iterations~\cite{dorigo_gambardella_acs_1997}. This sample-evaluate-reinforce cycle is a natural fit for heatmap decoding, because the heatmap supplies useful but noisy local evidence while the pheromone feedback can decide, over repeated sampling, which locally attractive edges are globally consistent, all without the per-instance tree search that makes MCTS expensive~\cite{xia_position_2024,pan_beyond_2025}. Among ACO variants, the Max--Min Ant System (\MMAS)~\cite{stutzle_maxmin_2000} is one of the strongest and most stable choices for the TSP, as it bounds the pheromone values between explicit limits and restarts on stagnation, which prevents premature convergence and makes the search stable at scale. We therefore design our decoder based on \MMAS. The central challenge is to incorporate the heatmap into the construction rule without displacing the pheromone feedback. A bias expressed in predictor-dependent units produces inconsistent effects across heatmaps, because the raw confidence magnitude varies across predictors. The bias should be calibrated so that the heatmap influences early construction and then pheromone feedback can later correct edges that the heatmap incorrectly favours, once instance-specific evidence accumulates.

This paper proposes \HeatACO{}, a scalable, predictor-agnostic heatmap-to-tour decoder for large-scale TSPs. \HeatACO{} adds a capped, degree-aware evidence bias to the \MMAS{} construction rule. The bias rewards only edges whose confidence exceeds a node's tour-degree capacity, while its scale is bounded by the pheromone dynamic range so that pheromone feedback can override misleading guidance. The evidence is folded into transition weights on fixed-size candidate lists, retaining $O(nk)$ per-ant sampling and parallel construction. Following established large-scale neural TSP protocols~\cite{fu_generalize_2021,pan_h-tsp_2023,sun_difusco_2023}, we evaluate fixed-heatmap decoding across the TSP500, TSP1K, and TSP10K benchmarks. Our contributions are as follows.

\begin{itemize}
\item We propose \HeatACO, a scalable, predictor-agnostic heatmap-to-tour decoder for large-scale TSPs. It extends \MMAS{} with a capped, degree-aware evidence bias whose strength is scaled automatically from the pheromone dynamic range. A single evidence configuration therefore decodes heatmaps from different pretrained predictors without retraining or predictor-specific tuning.

\item We evaluate \HeatACO{} against two commonly used fixed-heatmap decoders across large-scale TSPs. Across four heatmap sources, \HeatACO{} decodes the same fixed heatmaps to higher quality and in less time than the MCTS decoder, improving construction quality for every predictor on TSP500,  TSP1K, and TSP10K. It transfers well to out-of-distribution settings, including the directed Asymmetric \TSP{} (\ATSP{}).

\item We quantify the marginal value of heatmap guidance to \MMAS{} and identify two empirical boundaries: heatmap reliability under distribution shift and complementarity with geometric local search. Candidate-graph and confidence-band analyses distinguish failures associated with degraded heatmap evidence from cases in which reliable guidance adds little beyond strong geometric repair.

\end{itemize}
\section{Background}
\label{sec:background}

\subsection{Travelling Salesman Problem}
\label{subsec:tsp_definition}

The TSP is defined on a complete graph $G=(V,E)$ with node set $V$, $|V|=n$, and edge set $E$, where each edge $(i,j)\in E$ has a cost $d_{ij}$. A feasible solution is a closed tour that visits every node exactly once, written as a permutation $\pi=(\pi_1,\dots,\pi_n)$ of $V$ with tour length
\begin{equation}
\mathcal{L}(\pi)=\sum_{t=1}^{n} d_{\pi_t,\pi_{t+1}},\qquad \pi_{n+1}=\pi_1,
\label{eq:tour_length}
\end{equation}
and the goal is to find a tour of minimum length \cite{reinelt1991tsplib}. In the symmetric TSP, $d_{ij}=d_{ji}$ for every node pair, while in the \ATSP{}, $d_{ij}\neq d_{ji}$ for at least one pair \cite{jonker1987transforming}. We study both NP-hard variants~\cite{korte2011combinatorial}.

\subsection{Heatmap-Based Non-Autoregressive Neural Solvers}
\label{subsec:heatmap_nar}

A heatmap-based NAR solver comprises two stages: a neural heatmap predictor and a tour decoder. The predictor maps an instance representation $X$ to an edge-confidence matrix in one forward pass,
\begin{equation}
H = f_\theta(X) \in [0,1]^{n\times n},
\label{eq:heatmap_def}
\end{equation}
where each entry scores how likely the corresponding edge is to lie on a high-quality tour. A decoder then combines the heatmap with the distance matrix to produce $\hat{\pi}=g(H,D)=g(f_\theta(X),D)$; the complete solver returns a tour only after this second stage. Throughout this paper, \emph{predictor} refers to $f_\theta$, \emph{decoder} refers to $g$, and \emph{solver} refers to their composition. The predictors differ in architecture and training objective, but they share the same output interface: a dense confidence matrix over edges. For symmetric instances the heatmap is commonly symmetrised as $\bar{H}_{ij}=(H_{ij}+H_{ji})/2$~\cite{sun_difusco_2023}, while for the \ATSP{} the directed matrix is kept. We write $h_{ij}$ for the entry actually consumed by the decoder: $h_{ij}=\bar{H}_{ij}$ for symmetric TSP and TSPLIB instances, and $h_{ij}=H_{ij}$ for \ATSP. This common interface is what makes a predictor-agnostic decoder study possible. Because the confidence objective is defined edge by edge and does not encode the degree or subtour constraints of a feasible tour, the raw heatmap is not itself a tour and must be decoded.

\label{subsec:decoding_problem}

Heatmap-to-tour decoding is the second stage of a heatmap-based NAR solver and maps a fixed distance matrix $D$ and heatmap $H$ to a feasible tour $\hat{\pi}=g(H,D)$~\cite{fu_generalize_2021,pan_beyond_2025}. We hold $H$ fixed to study this stage independently and isolate the decoder's contribution to final solution quality. The decoder is a substantial algorithmic component: it extracts useful local evidence, rejects high-confidence false positives, keeps the tour feasible under the degree and subtour constraints, and runs fast enough not to erase the parallelism gained by NAR prediction.

\subsubsection{Greedy Merge}
\label{subsec:greedy_merge}

Greedy merge scores edges by a combined metric such as $s_{ij}=H_{ij}/d_{ij}$~\cite{sun_difusco_2023}, sorts them in descending order, and inserts them one by one subject to degree and subtour checks. The time cost is dominated by sorting the $O(n^2)$ edges, which takes $O(n^2\log n^2)$ time. Its appeal is speed and determinism, but it cannot revise an edge once accepted. A single misleading high-confidence edge can therefore force the rest of the construction toward a poor tour.

\subsubsection{MCTS-Guided $k$-opt Search}
\label{subsec:mcts_decoder}

These decoders use Monte Carlo Tree Search (MCTS)~\cite{fu_generalize_2021} to select sequences of $k$-opt moves, biasing proposals toward high-confidence heatmap edges. The per-round cost is $O(D_{\mathrm{MCTS}}B\,n(n+k))$ under the published formulation, and additional preprocessing for distance matrices, heatmaps, and candidate sets can add $O(n^2)$ to $O(n^2k)$ work. MCTS recovers higher-quality tours than greedy merge, but the quality it achieves is sensitive to the action space, the exploration constant, the rollout policy, and the search budget, and these must be tuned to each heatmap source and instance scale~\cite{pan_beyond_2025}. 

\vspace{-2mm}
\subsection{Related Work}
\label{subsec:related_work}


\emph{Ant colony optimisation for the TSP.} ACO is a constructive metaheuristic whose variants differ in how they manage the pheromone trail. Ant System~\cite{dorigo_maniezzo_colorni_ant_system_1996} introduced the basic sample-and-reinforce loop, in which all ants deposit pheromone in proportion to tour quality. Ant Colony System~\cite{dorigo_gambardella_acs_1997} added a pseudorandom proportional construction rule that sharpens exploitation and a local pheromone update that diversifies successive ants. \MMAS~\cite{stutzle_maxmin_2000} is the variant we build on: it lets only the best tour deposit, bounds every trail within an explicit interval $[\tau_{\min},\tau_{\max}]$, and reinitialises the trails on stagnation, which together prevent premature convergence and make the search reliable on large instances. The behaviour of these algorithms has been studied from convergence and invariance standpoints~\cite{dorigo_ant_2005,birattari_pellegrini_dorigo_invariance_2007}, and pairing ACO with strong local search remains competitive on routing problems~\cite{mavrovouniotis_aco_ls_dynamic_tsp_2017,tuani_haaco_3opt_tsp_2020}. \HeatACO{} leaves this \MMAS{} machinery unchanged and modifies only the construction weight, into which a fixed heatmap enters as a bounded evidence factor.

\emph{Neural combinatorial optimisation.} A second line learns solvers from data, surveyed in~\cite{bengio_machine_2021,mazyavkina_rl_co_survey_2021,cappart_gnn_co_reasoning_2023}. Autoregressive constructive solvers emit a tour one node at a time, from Pointer Networks~\cite{vinyals_pointer_2015} and reinforcement-learning policies~\cite{bello_neural_2017} to attention-based models~\cite{kool_attention_2019,bresson_transformer_2021}, while a complementary group learns to improve a given tour through learned local rewriting or test-time search~\cite{chen_learning_2019,hudson_graph_2022,hottung_efficient_2022}. Both exploit learned priors effectively, but rely on sequential decoding or repeated neural evaluation, which becomes the bottleneck at thousands of nodes. This bottleneck has motivated large-scale designs based on divide-and-conquer decomposition and transferable routing backbones~\cite{pan_h-tsp_2023,zheng_udc_2025,berto_routefinder_2024}, as well as tighter coupling of learning with classical local search~\cite{zhao_drl_local_search_vrp_2021}.

\emph{Heatmap-based NAR solvers and their decoders.} The setting of this paper is the family of NAR solvers that predict an edge-confidence heatmap in one forward pass and delegate feasibility to a separate decoder~\cite{joshi_efficient_2019,fu_generalize_2021,sun_difusco_2023}, which raises decoding to a first-order algorithmic problem. AttGCN~\cite{fu_generalize_2021} trains a graph convolutional network by supervised learning against optimal tours. DIMES~\cite{qiu_dimes_2022} learns a continuous edge parameterisation with a meta-learning reinforcement objective. UTSP~\cite{min_unsupervised_2023} uses an unsupervised surrogate loss that needs no labelled tours. DIFUSCO~\cite{sun_difusco_2023} casts edge prediction as a diffusion-based generative process. Despite their different training objectives, all four produce the same output in the form of a dense confidence matrix over edges, and therefore expose the same downstream interface that our decoder design exploits. UniCO~\cite{pan_unico_2024} extends this interface to the directed \ATSP{} and to non-Euclidean instances. Two commonly used fixed-heatmap decoders in this line of work are the greedy edge merging used by efficient diffusion and refinement pipelines~\cite{sun_difusco_2023,li_t2t_2023} and the MCTS-guided $k$-opt search of large-scale neural solvers~\cite{fu_generalize_2021,pan_beyond_2025}. The AttGCN evaluation on TSP10K still relies on MCTS, whose repeated per-instance search and sensitivity to heatmap source and scale mean that predictor scalability does not by itself resolve decoder scalability~\cite{fu_generalize_2021,pan_beyond_2025}. These limitations motivate a common decoder that can operate across fixed heatmaps and instance scales without retraining or decoder retuning. A separate group of recent solvers, including GLOP~\cite{ye_glop_2024}, GenSCO~\cite{liGenerationSearchOperator2025}, and MaskCO~\cite{chenMaskCOMaskedGeneration2025}, redesigns the generation or reconstruction mechanism itself, but are not predictor-agnostic decoders for an arbitrary fixed heatmap.

\emph{Neural components inside ACO.} The line closest to ours embeds learned components in the ACO loop. DeepACO~\cite{ye_deepaco_2023} trains a GNN to output the heuristic desirability that biases construction, GFACS~\cite{kim_ant_2024} learns a GFlowNet sampler to propose tours, and GTG-ACO~\cite{abirGTGACOGraph2025} jointly learns the heuristic and an initial pheromone field. These methods train a problem-specific neural model as part of the solver and couple it to a particular ACO configuration. \HeatACO{} differs in both inputs and goal: it consumes a heatmap that an external predictor has already produced, performs no training of its own, and is designed to decode heatmaps from \emph{different} predictors through one scale-free bias, so a new heatmap source needs no new network and no retuning. 
\section{The Proposed Method}
\label{sec:method}

\begin{figure}[!t]
  \centering
  \includegraphics[width=\columnwidth]{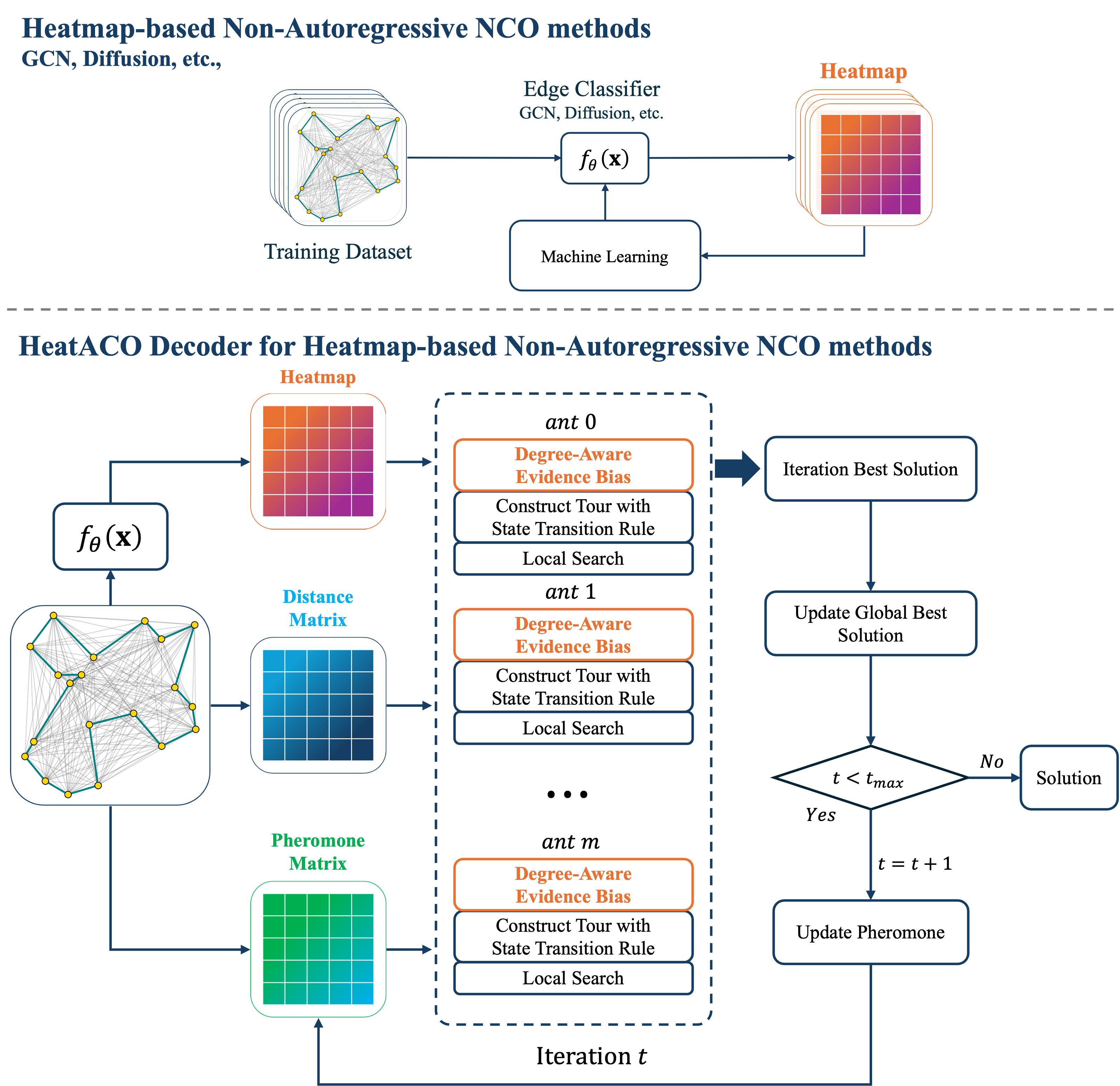}
  \caption{Overview of \HeatACO. Top: a heatmap-based non-autoregressive predictor (such as a GCN or diffusion model) is trained on a dataset and, at inference, predicts an edge-confidence heatmap $H$ for the instance. Bottom: the \HeatACO{} decoder takes the fixed heatmap together with the distance and pheromone matrices and runs a Max--Min Ant System. In each iteration, $m$ ants build tours in parallel, each turning the heatmap into a degree-aware evidence bias, constructing a tour by the state transition rule, and optionally applying local search. The iteration-best tour updates the global best and the pheromone matrix, and the loop repeats until the iteration budget $t_{\max}$ is reached, after which the best tour is returned.}
  \vspace{-4mm}
  \label{fig:pipeline_overview}
\end{figure}

\HeatACO{} takes a distance matrix $D$ and a precomputed heatmap $H$ as input and returns a feasible tour. It treats the predictor that produced $H$ as fixed, so it does not retrain the predictor, alter the heatmap, or learn a new neural model. Fig.~\ref{fig:pipeline_overview} gives an overview, from heatmap prediction by the fixed predictor to the iterative \HeatACO{} decoding loop.  Algorithm~\ref{alg:heataco} gives the full procedure.

\subsection{Tour Construction}
\label{subsec:construction}

Restricting construction to sparse candidate lists keeps the per-ant sampling cost at $O(nk)$ and keeps pheromone maintenance linear in the number of retained edges. Each node $i$ is given a construction candidate list $C_i\subset V\setminus\{i\}$ of fixed size $|C_i|=k=20$. We form $C_i$ from the heatmap-supported set
\begin{equation}
P_i=\{j\in V\setminus\{i\}: h_{ij}>\theta_h\},
\qquad \theta_h=10^{-4}.
\label{eq:heatmap_support}
\end{equation}
If $|P_i|>k$, we keep the $k$ nodes of $P_i$ closest to $i$, and if $|P_i|<k$, we fill the remaining slots with the nearest unused nodes by distance. Heatmap support admits useful non-local edges that a pure nearest-neighbour list would miss, while the geometric filling preserves connectivity when the heatmap is sparse.

An ant builds a tour by visiting nodes one at a time. From its current node $i$ it chooses the next node within the feasible neighbourhood $\mathcal{N}_i$, the candidate nodes in $C_i$ not yet visited, with probability
\begin{equation}
p_{ij} =
\frac{[\tau_{ij}]^{\alpha}\,[\eta_{ij}]^{\beta}\,[e_{ij}]^{\kappa}}
{\sum_{l\in\mathcal{N}_i}[\tau_{il}]^{\alpha}\,[\eta_{il}]^{\beta}\,[e_{il}]^{\kappa}},
\qquad j\in\mathcal{N}_i,
\label{eq:transition}
\end{equation}
where $\tau_{ij}$ is the pheromone trail, $\eta_{ij}=1/d_{ij}$ is the heuristic desirability, and $e_{ij}\ge 1$ is the heatmap evidence factor defined in Section~\ref{subsec:evidence}. The three construction terms therefore share the same multiplicative power-law form. When $\kappa=0$, or when an edge has $e_{ij}=1$, the heatmap multiplier is one and the transition weights recover the classical \MMAS{} form over the current candidate list. As in Ant Colony System~\cite{dorigo_gambardella_acs_1997}, the choice follows a pseudorandom proportional rule, in which with probability $q_0$ the ant takes the node with the largest numerator and otherwise samples from $p_{ij}$; we set $q_0=0$, so construction is purely probabilistic. When $\mathcal{N}_i$ is empty the ant moves to the nearest unvisited node by distance, which guarantees a feasible tour even where the candidates are exhausted. Algorithm~\ref{alg:ant_construction} summarises the construction.

\begin{algorithm}[!t]
\footnotesize
\caption{\HeatACO{} Decoding for a Fixed Heatmap}
\label{alg:heataco}
\begin{algorithmic}[1]
\REQUIRE Distance matrix $D$, heatmap $H$, $m$, $I$, $k$, $\alpha$, $\beta$, $\rho$, $c_{\mathrm{cap}}$
\ENSURE Best tour $T_{\mathrm{gb}}$
\STATE \textit{// Preparation, once per instance}
\STATE Build construction candidates $\{C_i\}$ from $H$ and $D$ \hfill $\triangleright$ \ref{subsec:construction}
\STATE Compute evidence $e_{ij}$ by~\eqref{eq:confidence}--\eqref{eq:evidence} and scale $\kappa$ by~\eqref{eq:q90}--\eqref{eq:kappa} \hfill $\triangleright$ \ref{subsec:evidence}
\STATE Construct NN tour $T_{\mathrm{NN}}$ (optionally one 2-opt pass)
\STATE Init $\tau_{\max},\tau_{\min}$ from $\mathcal{L}(T_{\mathrm{NN}})$ by~\eqref{eq:tau_init}; set $\tau\leftarrow\tau_{\max}$
\STATE $T_{\mathrm{gb}}\leftarrow T_{\mathrm{NN}}$
\STATE \textit{// Search}
\FOR{$t=1$ to $I$}
    \FOR{each of $m$ ants in parallel}
        \STATE Construct a tour by Algorithm~\ref{alg:ant_construction}; optionally apply local search \hfill $\triangleright$ \ref{subsec:local_search}
    \ENDFOR
    \STATE Identify iteration-best $T_{\mathrm{ib}}$; update restart-best $T_{\mathrm{rb}}$
    \IF{a new global best is found}
        \STATE Update $T_{\mathrm{gb}}$; recompute $\tau_{\max},\tau_{\min}$
    \ENDIF
    \STATE Evaporate pheromone on candidate edges
    \STATE Select elite tour $T^{+}$ by the $u_{\mathrm{gb}}$ schedule; deposit on $T^{+}$ by~\eqref{eq:pheromone_update}
    \IF{branching factor $<\mathrm{branch\_fac}$ (checked every $100$ iterations) and restart-best stagnation $>250$ iterations}
        \STATE Reset $\tau\leftarrow\tau_{\max}$ and restart-best $T_{\mathrm{rb}}$
    \ENDIF
    \STATE Update the $u_{\mathrm{gb}}$ schedule
\ENDFOR
\RETURN $T_{\mathrm{gb}}$
\end{algorithmic}
\end{algorithm}

\begin{algorithm}[!t]
\footnotesize
\caption{Ant tour construction}
\label{alg:ant_construction}
\begin{algorithmic}[1]
\REQUIRE $\tau$, $\eta$, $e$, $\kappa$, $\{C_i\}$, $q_0=0$
\ENSURE Feasible tour $T$
\STATE $T[0]\leftarrow$ random start node; mark visited
\FOR{$s=1$ to $n-1$}
    \STATE $i \leftarrow T[s-1]$; $\;\mathcal{N}_i \leftarrow C_i \setminus \{\text{visited}\}$
    \IF{$\mathcal{N}_i = \emptyset$}
        \STATE $j\leftarrow$ nearest unvisited node to $i$
    \ELSE
        \STATE $w_{ij} \leftarrow [\tau_{ij}]^{\alpha}[\eta_{ij}]^{\beta}[e_{ij}]^{\kappa}$ for $j\in\mathcal{N}_i$
        \STATE sample $j\in\mathcal{N}_i$ with probability $w_{ij}/\sum_{l}w_{il}$
    \ENDIF
    \STATE $T[s] \leftarrow j$; mark visited
\ENDFOR
\RETURN $T$
\end{algorithmic}
\end{algorithm}

\subsection{Pheromone Update}
\label{subsec:pheromone}

\HeatACO{} first constructs a nearest-neighbour tour $T_{\mathrm{NN}}$, applying one 2-opt pass when local search is enabled, and sets the pheromone bounds from its length $L_{\mathrm{nn}}=\mathcal{L}(T_{\mathrm{NN}})$ following the \MMAS{} convention~\cite{stutzle_maxmin_2000},
\begin{equation}
\tau_{\max}^{(0)} = \frac{1}{\rho\,L_{\mathrm{nn}}},\qquad
\tau_{\min}^{(0)} = \frac{\tau_{\max}^{(0)}}{2n},
\label{eq:tau_init}
\end{equation}
where $\rho$ is the evaporation rate. When local search is disabled we replace this lower bound with the standard \MMAS{} value derived from a target construction probability $p_{\mathrm{best}}=0.05$~\cite{stutzle_maxmin_2000}. Writing $p_x=\sqrt[n]{p_{\mathrm{best}}}$,
\begin{equation}
\tau_{\min}^{(0)} = \frac{\tau_{\max}^{(0)}\,(1-p_x)}{\tfrac{k-1}{2}\,p_x},
\label{eq:tau_min_nols}
\end{equation}
The pheromone matrix is then filled uniformly with $\tau_{\max}^{(0)}$. The heatmap is never written into the pheromone matrix, so the pheromone values remain the sole record of instance-specific search experience and the heatmap acts only through the construction step.

At the end of each iteration the pheromone evaporates and one elite tour $T^{+}$ deposits, with the bounded update
\begin{equation}
\tau_{ij}\leftarrow
\min\!\Big\{\tau_{\max},\,
\max\!\big\{\tau_{\min},\,
(1-\rho)\,\tau_{ij}+\Delta\tau_{ij}^{+}
\big\}\Big\},
\label{eq:pheromone_update}
\end{equation}
where the deposit on the elite tour is
\begin{equation}
\Delta\tau_{ij}^{+} =
\begin{cases}
1/\mathcal{L}(T^+), & (i,j)\in T^+,\\
0, & \text{otherwise}.
\end{cases}
\label{eq:pheromone_delta}
\end{equation}
Both bounds in~\eqref{eq:pheromone_update} are applied to the updated trail values. The elite tour follows the adaptive $u_{\mathrm{gb}}$ schedule of \MMAS~\cite{stutzle_maxmin_2000}, using the iteration-best tour on most iterations and the restart-best or global-best tour every $u_{\mathrm{gb}}$ iterations, with $u_{\mathrm{gb}}$ tightening from $25$ to $1$ as the search stalls. Whenever a new global-best tour is found, $\tau_{\max}$ and $\tau_{\min}$ are recomputed from its length. Every $100$ iterations the branching factor is checked, and if it falls below $\lambda_{\mathrm{bf}}=1+10^{-5}$ while the restart-best tour has not improved for more than $250$ iterations, the pheromone is reset to $\tau_{\max}$ and the restart-best tour is cleared, while the global-best tour is kept.


\subsection{Degree-Aware Evidence Bias}
\label{subsec:evidence}

The single heatmap-dependent component is the evidence factor $e_{ij}$ and its scale $\kappa$ (we use $\kappa$ rather than $\lambda$, which is reserved for the \MMAS{} branching-factor threshold $\lambda_{\mathrm{bf}}$ in Section~\ref{subsec:pheromone}). We read the heatmap as ordinal local evidence rather than as a calibrated probability, and use the thresholded values directly to avoid predictor-specific calibration. Let $d^\star$ be the local degree target, with $d^\star=2$ for symmetric TSP and TSPLIB and $d^\star=1$ for the outgoing decision in \ATSP. For each candidate edge $j\in C_i$ we take the thresholded confidence
\begin{equation}
c_{ij} = \begin{cases}
h_{ij}, & h_{ij} > \theta_h,\\
0, & \text{otherwise},
\end{cases}
\label{eq:confidence}
\end{equation}
with confidence floor $\varepsilon = 10^{-9}$ used to avoid zero ratios. Sorting the positive values in row $i$ as $c_i^{(1)}\ge c_i^{(2)}\ge\cdots$ and setting missing order statistics to zero, the degree baseline is the first confidence level the node cannot use within its tour degree,
\begin{equation}
b_i = \max\!\big(c_i^{(d^\star+1)},\,\varepsilon\big).
\label{eq:background}
\end{equation}
For symmetric TSP ($d^\star=2$) this is the third-highest confidence in the row, because a node accepts two incident tour edges and any confidence beyond the second-best edge lies above the baseline, and for \ATSP{} ($d^\star=1$) it is the second-highest confidence, since a row represents one outgoing edge. The evidence factor is the confidence ratio above this baseline, lower-bounded by one,
\begin{equation}
e_{ij} = \max\!\left(\frac{\max(c_{ij},\varepsilon)}{b_i},\;1\right).
\label{eq:evidence}
\end{equation}
An edge therefore receives additional evidence, $e_{ij}>1$, only when its confidence exceeds the degree baseline, i.e., when the heatmap distinguishes it beyond the node's tour-degree capacity. When a row has many similar scores the ratios approach one, and the decoder falls back to distance and pheromone. The local search uses independent geometric candidate lists, so the evidence affects only construction and never the repair operator.

The scale $\kappa$ should let the heatmap bias early sampling without permanently overriding the pheromone feedback, so we cap it by the pheromone dynamic range. With $q_{0.90}$ the 90th percentile of the positive log-margins,
\begin{equation}
q_{0.90} = Q_{0.90}\!\left(\{\log e_{ij} : i\in V,\; j\in C_i,\; e_{ij}>1\}\right),
\label{eq:q90}
\end{equation}
and $\widehat{R}_{\tau}$ the active pheromone dynamic range,
\begin{equation}
\widehat{R}_{\tau} = \frac{\tau_{\max}}{\tau_{\min}},
\label{eq:tau_ratio}
\end{equation}
the scale is
\begin{equation}
\kappa =
\begin{cases}
\dfrac{c_{\mathrm{cap}}\,\alpha\log \widehat{R}_{\tau}}{q_{0.90}},
& q_{0.90}>0,\ \widehat{R}_{\tau}>1,\\[2pt]
0, & \text{otherwise},
\end{cases}
\label{eq:kappa}
\end{equation}
with $c_{\mathrm{cap}}=0.5$. At the 90th percentile the log of the evidence multiplier is then $\kappa q_{0.90}=c_{\mathrm{cap}}\alpha\log\widehat{R}_{\tau}$, so under our setting $\alpha=1$ it equals half the log pheromone range; for general $\alpha$ it is $(\alpha/2)\log\widehat{R}_{\tau}$. The heatmap thus shapes early sampling while the pheromone feedback keeps enough dynamic range to correct the bias over time. If the heatmap provides no edge with $e_{ij}>1$, then $\kappa=0$ and the transition weights take the classical \MMAS{} form over the current candidate list. The rule uses only the heatmap scores, the candidate graph, and the \MMAS{} configuration, and it does not use known tours or per-predictor tuning. For efficiency the implementation folds $[e_{ij}]^{\kappa}$ into the heuristic term during preparation and clips the log-weight $\kappa\log e_{ij}$ at a large constant to avoid overflow, which leaves~\eqref{eq:transition} unchanged.

\subsection{Local Search}
\label{subsec:local_search}

A constructed tour can optionally be refined by local search before it updates the pheromone. For symmetric TSP and TSPLIB we use first-improvement 2-opt~\cite{croes_2opt_1958} and 3-opt, both restricted to geometric nearest-neighbour candidate lists of size $k_{\mathrm{ls}}=20$ with don't-look bits, and for \ATSP{} we use a directed 2-opt based on the Jonker--Volgenant reduction~\cite{jonker1987transforming}. The local search runs on geometric candidates and is independent of the heatmap-supported construction lists, so the heatmap guides construction while the repair operator remains purely distance-based. Improved tours feed the pheromone update, so the benefit of local search propagates to later iterations through the pheromone distribution.
\section{Experimental Design}
\label{sec:experimental_design}

\subsection{Datasets}
\label{subsec:datasets}

All benchmark instances are drawn from ML4CO-Bench-101~\cite{maML4COBench101BenchmarkMachine2025}. The in-distribution evaluation uses its TSP500, TSP1K, and TSP10K suites, which are the uniform Euclidean benchmarks originally released with AttGCN~\cite{fu_generalize_2021}. TSP500 and TSP1K each contain 128 instances, and TSP10K contains 16 instances. For TSP500 and TSP1K, $L^\star_i$ is the Concorde optimum~\cite{applegate_tsp_computational_study_2006}; for TSP10K, $L^\star_i$ is the known optimum supplied with the benchmark. The four heatmaps for these instances come from the publicly released pretrained checkpoints of AttGCN~\cite{fu_generalize_2021}, DIMES \cite{qiu_dimes_2022}, UTSP \cite{min_unsupervised_2023}, and DIFUSCO \cite{sun_difusco_2023}, with no retraining or fine-tuning.

Generalisation is evaluated under several qualitatively distinct shifts, all drawn from the same benchmark. Clustered TSP500 and Gaussian TSP500 (128 instances each) test spatial distribution shift at fixed scale. Three TSPLIB~\cite{reinelt1991tsplib} instances, pcb442, pr1002, and pr2392, test structural shift from synthetic uniform coordinates to application-derived geometries. GenSCO heatmaps~\cite{liGenerationSearchOperator2025} from the released $C=160$ configuration test transfer to a structurally different heatmap source. MatDIFFNet heatmaps~\cite{pan_unico_2024} for ATSP50 and ATSP100 test transfer to directed instances.

\subsection{Compared Methods}
\label{subsec:compared_methods}

Our single controlled baseline is standard \MMAS. It uses the same parameters, candidate-list sizes, local-search settings, hardware, and random seeds as \HeatACO. This matched comparison isolates the marginal gain of the heatmap, and it is the only one we use for paired statistical testing. All other methods serve as external reference points and are reported as published. They include the classical solvers Concorde~\cite{applegate_tsp_computational_study_2006} and LKH-3~\cite{helsgaun_extension_2017}, the NCO solver LEHD~\cite{luo_neural_2023}, and the neural-ACO solvers DeepACO~\cite{ye_deepaco_2023} and GFACS~\cite{kim_ant_2024}. Two commonly used fixed-heatmap decoders are our direct points of comparison: greedy edge merging, introduced as the fast decoder of DIFUSCO~\cite{sun_difusco_2023}, and MCTS-guided $k$-opt search, introduced with AttGCN~\cite{fu_generalize_2021}. For the MCTS decoder, we use the results of Pan et al.~\cite{pan_beyond_2025}, who tuned its hyperparameters per heatmap to obtain its best published quality. For each heatmap source we evaluate \HeatACO{} (i.e., without local-search refinement), \HeatACO{}+2-opt, and \HeatACO{}+3-opt on symmetric TSP and TSPLIB, and \HeatACO{} and \HeatACO{}+2-opt on \ATSP.

These external references were obtained on stronger hardware than ours. They are reported on 16- to 128-core CPUs together with high-performance GPUs for the neural inference, whereas all \HeatACO{} and \MMAS{} results use 8 CPU cores and no GPU. The comparison is therefore conservative for \HeatACO: where it matches or improves on a reference, it does so from a weaker computational budget.

\subsection{Metrics}
\label{subsec:metrics}

The primary quality metric is the optimality gap,
\begin{equation}
\mathrm{Gap}_{i,s}(\%)=\frac{\mathcal{L}(T_{i,s})-L^\star_i}{L^\star_i}\times 100.
\label{eq:gap}
\end{equation}
Here \(T_{i,s}\) is the best tour found for instance \(i\) under seed \(s\), and \(L^\star_i\) is that instance's reference length. For benchmark suites containing multiple instances, we first average the 30 seed results for each instance, and then report the mean and standard deviation of these instance-level means. For individual TSPLIB instances, the mean and standard deviation are computed across the 30 seeds. Where reported, runtime is given as per-instance wall-clock decoding time, which is the metric available for the published external baselines. For heatmap-based methods we list heatmap inference time separately from decoding time, and a leading ``$+$'' denotes the decoding time added after heatmap inference.

All statistical tests compare \HeatACO{} against the matched standard \MMAS{} baseline only. For benchmark suites containing multiple instances, we average the gap over the 30 seeds for each instance and apply a two-sided paired Wilcoxon signed-rank test~\cite{derrac_nonparametric_tests_2011} across instances at $p<0.05$; for individual TSPLIB instances, the test is applied across the 30 paired seeds. A significant result is marked as an improvement if the corresponding mean gap is lower for \HeatACO{} and as a degradation otherwise. The win, tie, and loss counts for multi-instance suites are computed from the same per-instance means.

\subsection{Parameter Settings}
\label{subsec:parameter_settings}

\HeatACO{} and \MMAS{} share the same core parameters: $m=32$ ants running in parallel, each with its own random number generator, with a barrier ensuring that all ants finish before the pheromone update; $I=5000$ iterations; construction candidate list size $k=20$; local-search candidate list size $k_{\mathrm{ls}}=20$; $\alpha=1$; $\beta=2$; $\rho=0.8$ for symmetric TSP/TSPLIB and $\rho=0.2$ for \ATSP; and $q_0=0$ for pure stochastic sampling. The heatmap threshold is $\theta_h=10^{-4}$, the confidence floor is $\varepsilon=10^{-9}$, the evidence quantile is $Q_{0.90}$, and the cap factor is $c_{\mathrm{cap}}=0.5$. Branching-factor monitoring is performed every $100$ iterations with band fraction $\xi_{\mathrm{bf}}=0.05$ and branching threshold $\lambda_{\mathrm{bf}}=1+10^{-5}$; restarts are triggered after more than $250$ consecutive iterations without restart-best improvement.

Every experimental setting is evaluated over 30 independent random seeds. The decoder is implemented in C++ and all \HeatACO{} and \MMAS{} experiments use 8 CPU cores. The cap factor $c_{\mathrm{cap}}=0.5$ was fixed before the final 30-run evaluation and was not selected by tuning on the benchmark results, and Section~\ref{subsec:parameter_sensitivity} reports its sensitivity.
\section{Main Results}
\label{sec:main_results}

\begin{table*}[!htbp]
  \centering
  \caption{Main Results on TSP500, TSP1K, and TSP10K.}
  \label{tab:main}
  \TableFont
  \setlength{\tabcolsep}{3.2pt}
  \begin{tabular}{@{}cc|cc|cc|cc@{}}
    \hline
    \multirow{2}{*}{\textbf{Method}} & \multirow{2}{*}{\textbf{Type}} & \multicolumn{2}{c|}{\textbf{TSP500}} & \multicolumn{2}{c|}{\textbf{TSP1K}} & \multicolumn{2}{c}{\textbf{TSP10K}} \\
    & & Gap $\downarrow$ & Time $\downarrow$ & Gap $\downarrow$ & Time $\downarrow$ & Gap $\downarrow$ & Time $\downarrow$ \\
    \hline
    Concorde~\cite{applegate2006concorde} & Exact & --- & 17.65\,s & --- & 3.12\,m & --- & --- \\
    LKH-3~\cite{helsgaun_extension_2017} & Heuristic & --- & 21.69\,s & --- & 1.2\,m & --- & 4.13\,m \\
    MCTS~\cite{pan_beyond_2025} & MCTS & 0.66 & 50\,s & 1.16 & 100\,s & 3.79 & 16.67\,m \\
    \hline
    \multirow{3}[0]{*}{MMAS \cite{stutzle_maxmin_2000}} & MMAS & 6.14{\footnotesize\textcolor{gray}{$\pm$0.30}} & 2.28\,s & 7.73{\footnotesize\textcolor{gray}{$\pm$0.26}} & 5.29\,s & 38.32{\footnotesize\textcolor{gray}{$\pm$0.30}} & 7.61\,m \\
     & MMAS+2-opt & 0.16{\footnotesize\textcolor{gray}{$\pm$0.05}} & 4.25\,s & 0.42{\footnotesize\textcolor{gray}{$\pm$0.05}} & 9.13\,s & 1.19{\footnotesize\textcolor{gray}{$\pm$0.03}} & 2.06\,m \\
     & MMAS+3-opt & 0.0045{\footnotesize\textcolor{gray}{$\pm$0.0060}} & 14.01\,s & 0.06{\footnotesize\textcolor{gray}{$\pm$0.02}} & 29.64\,s & 0.41{\footnotesize\textcolor{gray}{$\pm$0.02}} & 9.22\,m \\
    \hline
    DeepACO~\cite{ye_deepaco_2023} & NAR+ACO & 1.84 & 10\,s & 3.16 & 32\,s & --- & --- \\
    GFACS~\cite{kim_ant_2024} & NAR+ACO & 1.56 & 15\,s & 2.63 & 66\,s & --- & --- \\
    \multirow{2}{*}{LEHD~\cite{luo_neural_2023}} & AR+Greedy & 1.56 & 0.14\,s & 3.17 & 0.75\,s & --- & --- \\
     & +Partial Re-Construction 1000 & 0.17 & 33.75\,s & 0.72 & 3.28\,m & --- & --- \\
    GenSCO ($C{=}40$)~\cite{liGenerationSearchOperator2025} & NAR & 0.016 & 18\,s & 0.26 & 46\,s & --- & --- \\
    GenSCO ($C{=}160$)~\cite{liGenerationSearchOperator2025} & NAR+2-opt & 0.01 & 1.2\,m & 0.04 & 3.0\,m & --- & --- \\
    \hline
    AttGCN \cite{fu_generalize_2021} & \multirow{4}{*}{NAR+Greedy} & 47.78 & 0.2\,s & 65.52 & 0.34\,s & 184.90 & 15.6\,s \\
    DIMES \cite{qiu_dimes_2022} & & 76.28 & 0.45\,s & 101.61 & 0.98\,s & 359.15 & 17.43\,s \\
    UTSP \cite{min_unsupervised_2023} & & 32.28 & 0.64\,s & 44.94 & 1.57\,s & --- & --- \\
    DIFUSCO \cite{sun_difusco_2023} & & 10.85 & 1.69\,s & 13.06 & 5.56\,s & 36.75 & 1.78\,m \\
    \hline
    AttGCN & \multirow{4}{*}{+MCTS~\cite{pan_beyond_2025}} & 0.69 & 0.2\,s+50\,s & 1.09 & 0.34\,s+1.67\,m & 3.02 & 15.6\,s+16.67\,m \\
    DIMES & & 0.69 & 0.98\,s+50\,s & 1.11 & 0.98\,s+1.67\,m & 3.05 & 17.43\,s+16.67\,m \\
    UTSP & & 0.90 & 0.64\,s+50\,s & 1.53 & 1.57\,s+1.67\,m & --- & --- \\
    DIFUSCO & & 0.51 & 1.69\,s+50\,s & 0.53 & 5.56\,s+1.67\,m & 2.35 & 1.78\,m+16.67\,m \\
    \hline
    AttGCN & \multirow{4}{*}{+\HeatACO{}} & 3.60{\footnotesize\textcolor{gray}{$\pm$0.31}} & 0.2\,s+1.95\,s & 4.85{\footnotesize\textcolor{gray}{$\pm$0.31}} & 0.34\,s+4.58\,s & 23.06{\footnotesize\textcolor{gray}{$\pm$0.25}} & 15.6\,s+3.70\,m \\
    DIMES & & 4.65{\footnotesize\textcolor{gray}{$\pm$0.29}} & 0.98\,s+2.01\,s & 5.91{\footnotesize\textcolor{gray}{$\pm$0.21}} & 0.98\,s+4.32\,s & 27.36{\footnotesize\textcolor{gray}{$\pm$0.28}} & 17.43\,s+3.50\,m \\
    UTSP & & 4.37{\footnotesize\textcolor{gray}{$\pm$0.28}} & 0.64\,s+2.01\,s & 5.75{\footnotesize\textcolor{gray}{$\pm$0.25}} & 1.57\,s+4.18\,s & --- & --- \\
    DIFUSCO & & 0.81{\footnotesize\textcolor{gray}{$\pm$0.37}} & 1.69\,s+1.71\,s & 1.07{\footnotesize\textcolor{gray}{$\pm$0.38}} & 5.56\,s+4.28\,s & 9.20{\footnotesize\textcolor{gray}{$\pm$0.42}} & 1.78\,m+2.52\,m \\
    \hline
    AttGCN & \multirow{4}{*}{+\HeatACO{}+2-opt} & 0.14{\footnotesize\textcolor{gray}{$\pm$0.06}} & 0.2\,s+4.08\,s & 0.39{\footnotesize\textcolor{gray}{$\pm$0.06}} & 0.34\,s+8.49\,s & 1.75{\footnotesize\textcolor{gray}{$\pm$0.12}} & 15.6\,s+2.16\,m \\
    DIMES & & 0.15{\footnotesize\textcolor{gray}{$\pm$0.05}} & 0.98\,s+4.09\,s & 0.40{\footnotesize\textcolor{gray}{$\pm$0.05}} & 0.98\,s+8.40\,s & 1.29{\footnotesize\textcolor{gray}{$\pm$0.04}} & 17.43\,s+1.87\,m \\
    UTSP & & 0.14{\footnotesize\textcolor{gray}{$\pm$0.05}} & 0.64\,s+4.21\,s & 0.40{\footnotesize\textcolor{gray}{$\pm$0.05}} & 1.57\,s+8.47\,s & --- & --- \\
    DIFUSCO & & 0.12{\footnotesize\textcolor{gray}{$\pm$0.08}} & 1.69\,s+3.99\,s & 0.20{\footnotesize\textcolor{gray}{$\pm$0.07}} & 5.56\,s+8.37\,s & 1.49{\footnotesize\textcolor{gray}{$\pm$0.10}} & 1.78\,m+2.37\,m \\
    \hline
    AttGCN & \multirow{4}{*}{+\HeatACO{}+3-opt} & 0.0036{\footnotesize\textcolor{gray}{$\pm$0.0054}} & 0.2\,s+14.82\,s & 0.05{\footnotesize\textcolor{gray}{$\pm$0.02}} & 0.34\,s+28.54\,s & 0.73{\footnotesize\textcolor{gray}{$\pm$0.09}} & 15.6\,s+8.14\,m \\
    DIMES & & 0.0037{\footnotesize\textcolor{gray}{$\pm$0.0052}} & 0.98\,s+14.23\,s & 0.06{\footnotesize\textcolor{gray}{$\pm$0.02}} & 0.98\,s+27.01\,s & 0.44{\footnotesize\textcolor{gray}{$\pm$0.03}} & 17.43\,s+6.65\,m \\
    UTSP & & 0.0040{\footnotesize\textcolor{gray}{$\pm$0.0054}} & 0.64\,s+13.63\,s & 0.05{\footnotesize\textcolor{gray}{$\pm$0.02}} & 1.57\,s+27.10\,s & --- & --- \\
    DIFUSCO & & 0.01{\footnotesize\textcolor{gray}{$\pm$0.02}} & 1.69\,s+14.32\,s & 0.04{\footnotesize\textcolor{gray}{$\pm$0.03}} & 5.56\,s+26.43\,s & 0.81{\footnotesize\textcolor{gray}{$\pm$0.04}} & 1.78\,m+7.25\,m \\
    \hline
  \end{tabular}

  \vspace{2pt}
  \begin{minipage}{\textwidth}
  \TableNoteFont\emph{Note:} Gap cells report mean\textcolor{gray}{$\pm$std} for \MMAS{} and \HeatACO{} rows; std is computed over instance-level mean gaps after averaging the 30 seeds. Other neural-decoder rows are published reference results. Time is per-instance wall-clock seconds (s) or minutes (m). Dashes in reference-solver gap cells denote the defining reference gap, and other dashes denote unavailable published results. `+${}$' in time entries denotes additional decoding time beyond fixed heatmap inference. External references use 16--128 CPU cores with a high-performance GPU; \HeatACO{} and \MMAS{} use 8 CPU cores and no GPU. Lower values are better for both gap and time.
  \end{minipage}
  \vspace{-4mm}
\end{table*}


This section separates two comparisons. First, against published MCTS results for the same fixed heatmaps, \HeatACO{}+2-opt obtains lower gaps and shorter reported decoding times in every available in-distribution comparison from TSP500 to TSP10K. Second, against matched standard \MMAS{} baselines, heatmap guidance improves construction at every tested scale and remains beneficial in most local-search settings on TSP500 and TSP1K and in the directed \ATSP, while its contribution becomes negative after local search on TSP10K or under some severe distribution shifts.

\vspace{-1mm}
\subsection{In-Distribution Results}
\label{subsec:in_distribution}

The fixed heatmaps carry useful edge information, but greedy decoding exploits only a small fraction of this information. Across the four heatmap sources, as shown in Table~\ref{tab:main}, the greedy gaps range from $10.85\%$ to $76.28\%$ on TSP500 and from $13.06\%$ to $101.61\%$ on TSP1K, and they remain between $36.75\%$ and $359.15\%$ on TSP10K. These values are far above the standard \MMAS{} baselines, which shows that a heatmap alone does not yield a good tour. Greedy merging cannot revise an early edge, so a few high-confidence false positives commit the construction to a poor completion. 

Before any local search, the evidence bias already turns \MMAS{} construction from a weak sampler into a strong one. On TSP500 it lowers the mean gap from $6.14\%$ for standard \MMAS{} to $0.81\%$ with the DIFUSCO heatmap and to $3.60\%$--$4.65\%$ with the other three, and on TSP1K it lowers the gap from $7.73\%$ to $1.07\%$ with DIFUSCO and to $4.85\%$--$5.91\%$ with the others. The gain persists at TSP10K: against the $38.32\%$ \MMAS{} gap, the three available heatmaps reach $9.20\%$--$27.36\%$. As Table~\ref{tab:mmas_significance} shows, construction-only \HeatACO{} is significantly better on every instance at every available scale and for every predictor. This construction-only setting isolates the contribution of the heatmap before local search affects the final tour, and it shows that the bias moves the early sampling distribution into promising regions of the search space that standard \MMAS{} reaches only after many more iterations. The gap between DIFUSCO and the other three predictors is large and consistent, and it tracks how cleanly each heatmap separates tour edges from the rest. We quantify this separation in Section~\ref{sec:further_analysis}, where the precision of the edges with additional evidence is highest for DIFUSCO, which is the same ordering seen here.

With 2-opt, \HeatACO{} decodes a fixed heatmap to higher quality and in less time than the MCTS decoder. On TSP500 the four heatmaps reach $0.12\%$--$0.15\%$ with \HeatACO{}+2-opt, against $0.16\%$ for \MMAS{}+2-opt and $0.51\%$--$0.90\%$ for the same heatmaps decoded by MCTS, and on TSP1K \HeatACO{}+2-opt reaches $0.20\%$--$0.40\%$ against $0.42\%$ for \MMAS{}+2-opt and $0.53\%$--$1.53\%$ for MCTS. The scope of this claim is narrow: stronger generation and reconstruction solvers such as GenSCO and LEHD remain competitive at this scale, so the comparison is between \HeatACO{} and the greedy / MCTS decoders for the same heatmap family rather than against all neural solvers.

We include 3-opt as the strongest local-search configuration, since it is the setting in which geometric repair is strongest and therefore the most demanding test of whether the heatmap still adds value. On TSP500 and TSP1K, \HeatACO{}+3-opt reaches gaps below 0.1\% ($0.0036\%$--$0.01\%$ and $0.04\%$--$0.06\%$). The marginal contribution of the heatmap diminishes as the local search strengthens, because the bias mainly supplies structure that 2-opt and 3-opt would otherwise have to recover. On TSP1K, \HeatACO{}+3-opt is still significantly better than \MMAS{}+3-opt for all four predictors (Table~\ref{tab:mmas_significance}), so the heatmap remains useful even under the strongest local search at this scale. On TSP500, the smallest scale, the heatmap's marginal value vanishes for the most reliable heatmap: DIFUSCO+\HeatACO{}+3-opt is significantly worse than \MMAS{}+3-opt, at $0.01\%$ against $0.0045\%$. This is the same mechanism as the large-scale reversal, but reached from the opposite extreme: once 3-opt resolves the tour structure on a small instance, the heatmap has no further structure to contribute.

At TSP10K, the construction and local-search regimes separate sharply. Without local search, all three available heatmaps improve over \MMAS{} on all 16 instances. With 2-opt, \HeatACO{} still improves on MCTS for every heatmap, for instance $1.29\%$ for DIMES+\HeatACO{}+2-opt against $3.05\%$ for DIMES+MCTS, but it no longer outperforms the matched \MMAS{}+2-opt baseline at $1.19\%$. The same holds under 3-opt, where DIMES+\HeatACO{}+3-opt reaches $0.44\%$ against $0.41\%$ for \MMAS{}+3-opt. Section~\ref{sec:further_analysis} shows that the TSP10K heatmaps retain sparse, high-coverage edge evidence; the reversal therefore marks reduced complementarity with strong geometric repair, rather than a failure of the heatmap itself.

These quality changes are reached at low decoding cost relative to the baseline search. The heatmap restricts construction to a sparse candidate list, so heatmap-guided decoding incurs no systematic runtime overhead. On TSP1K the decoding time of \HeatACO{} is $4.2$--$4.6$\,s without local search and $8.4$--$8.5$\,s with 2-opt, comparable to the $9.1$\,s of \MMAS{}+2-opt; construction-only \HeatACO{} thus takes $4.2$--$4.6$\,s while bringing the DIFUSCO gap to $1.07\%$. At TSP10K, construction-only \HeatACO{} takes $2.52$--$3.70$\,m versus $7.61$\,m for \MMAS{}, while the 2-opt and 3-opt variants remain comparable to or faster than their matched baselines. Against the MCTS decoder the difference is large: DIFUSCO+\HeatACO{}+2-opt decodes in $8.4$\,s on TSP1K against the $100$\,s of MCTS, with a lower gap.

\begin{table}[!t]
  \centering
  \caption{Paired Wins/Ties/Losses Against Matched \MMAS{} Variants.}
  \label{tab:mmas_significance}
  \TableFont
  \setlength{\tabcolsep}{2.6pt}
  \begin{tabular}{@{}cc|ccc@{}}
    \hline
    \textbf{Heatmap} & \textbf{Decoder} & \textbf{TSP500} & \textbf{TSP1K} & \textbf{TSP10K} \\
    \hline
    AttGCN & \multirow{4}{*}{\HeatACO} & 128/0/0\SigBetter & 128/0/0\SigBetter & 16/0/0\SigBetter \\
    DIMES & & 128/0/0\SigBetter & 128/0/0\SigBetter & 16/0/0\SigBetter \\
    UTSP & & 128/0/0\SigBetter & 128/0/0\SigBetter & --- \\
    DIFUSCO & & 128/0/0\SigBetter & 128/0/0\SigBetter & 16/0/0\SigBetter \\
    \hline
    AttGCN & \multirow{4}{*}{\HeatACO{}+2-opt} & 99/0/29\SigBetter & 91/0/37\SigBetter & 0/0/16\SigWorse \\
    DIMES & & 76/0/52\SigBetter & 74/0/54\SigBetter & 0/0/16\SigWorse \\
    UTSP & & 90/0/38\SigBetter & 84/0/44\SigBetter & --- \\
    DIFUSCO & & 84/0/44\SigBetter & 128/0/0\SigBetter & 0/0/16\SigWorse \\
    \hline
    AttGCN & \multirow{4}{*}{\HeatACO{}+3-opt} & 52/43/33\SigBetter & 84/0/44\SigBetter & 0/0/16\SigWorse \\
    DIMES & & 52/41/35\SigBetter & 71/0/57\SigBetter & 2/0/14\SigWorse \\
    UTSP & & 41/42/45\SigTie & 80/0/48\SigBetter & --- \\
    DIFUSCO & & 36/32/60\SigWorse & 102/0/26\SigBetter & 0/0/16\SigWorse \\
    \hline
  \end{tabular}

  \vspace{2pt}
  \begin{minipage}{\columnwidth}
  \TableNoteFont\emph{Note:} Each cell reports W/T/L = wins/ties/losses against the matched \MMAS{} variant, computed over per-instance mean gaps after averaging the 30 seeds. The arrow combines a two-sided paired Wilcoxon signed-rank test ($p<0.05$) with the sign of the mean gap delta, so \SigBetter{} means significant with the mean favouring \HeatACO, \SigWorse{} means significant with the mean favouring \MMAS, and \SigTie{} means not significant.
  \end{minipage}
  \vspace{-2mm}
\end{table}

On TSP1K the win count decreases monotonically as the local search strengthens: $128/0/0$ without local search, a clear majority under 2-opt, a smaller but still significant majority under 3-opt, as stronger geometric repair recovers more of what the heatmap would otherwise supply. Across instance scales, construction-only guidance remains uniformly effective through TSP10K, where all three predictors win all 16 instances, but the same scale increase produces a reversal after local search: every 2-opt row loses all 16 instances and the 3-opt rows lose at least 14.

The agreement between W/T/L counts, signed mean differences, and Wilcoxon directions rules out a tail-driven explanation at TSP10K. The heatmap is consistently useful before local search and consistently non-complementary after 2-opt; under 3-opt, DIMES is close in mean gap but still loses on 14 of 16 instances.

\begin{figure}[!t]
  \centering
  \includegraphics[width=\columnwidth]{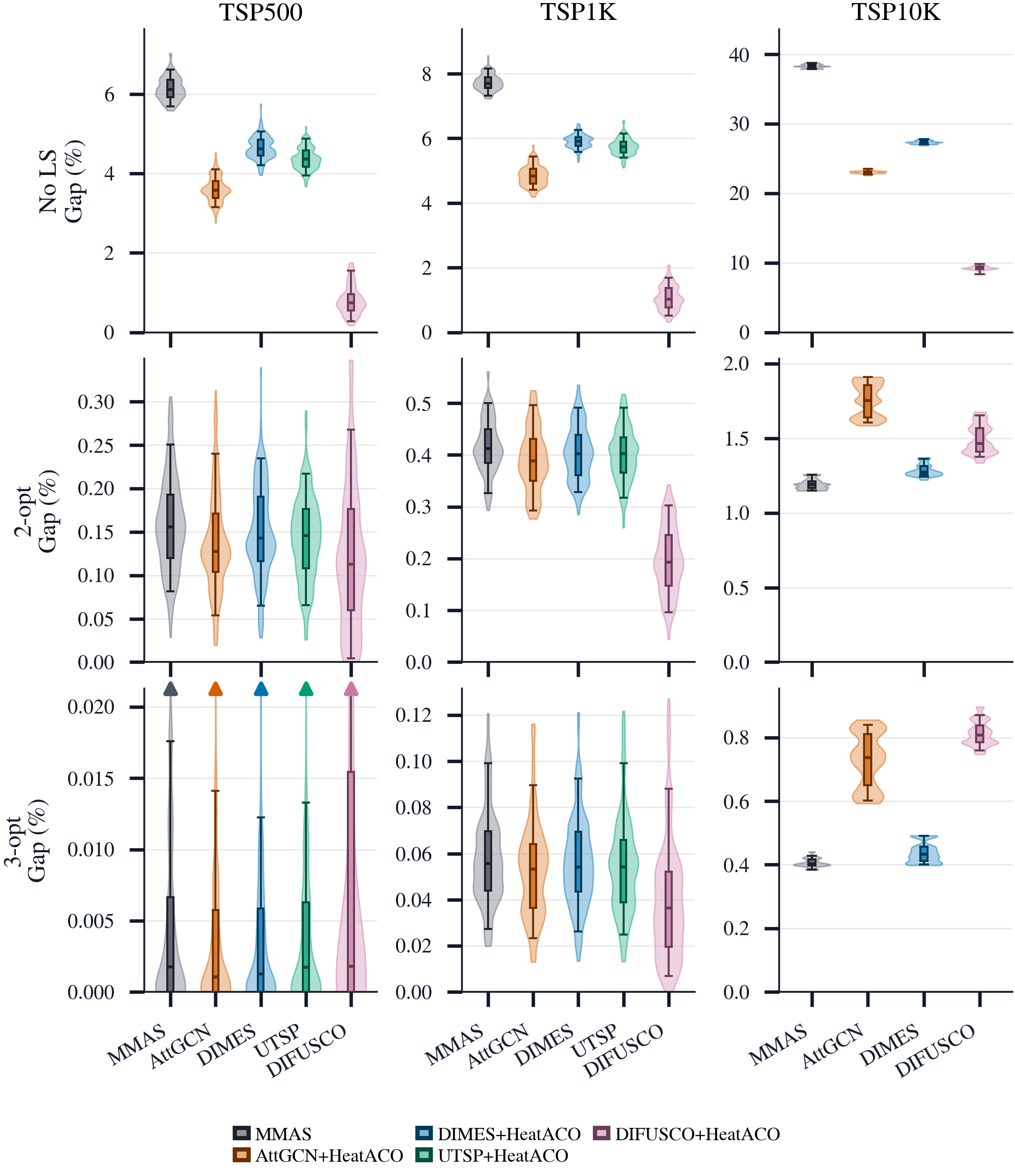}
  \caption{Instance-level gap distributions for the main in-distribution benchmarks. Columns correspond to TSP scale and rows to the local-search setting. Each violin is formed from instance-level mean gaps after averaging the 30 independent runs for each instance.}
  \vspace{-4mm}
  \label{fig:main_gap_violin}
\end{figure}

Fig.~\ref{fig:main_gap_violin} visualises the same controlled comparison at the distribution level. Without local search, every heatmap-guided distribution shifts downward relative to matched \MMAS{} at all available scales. On TSP500 and TSP1K this advantage usually persists under 2-opt, while with 3-opt the distributions become narrow and close because both decoders are near optimal. On TSP10K with local search, the \MMAS{} distribution is lower than the heatmap-guided variants, matching the boundary identified by the W/T/L analysis.

\begin{figure}[!h]
  \centering
  \includegraphics[width=\columnwidth]{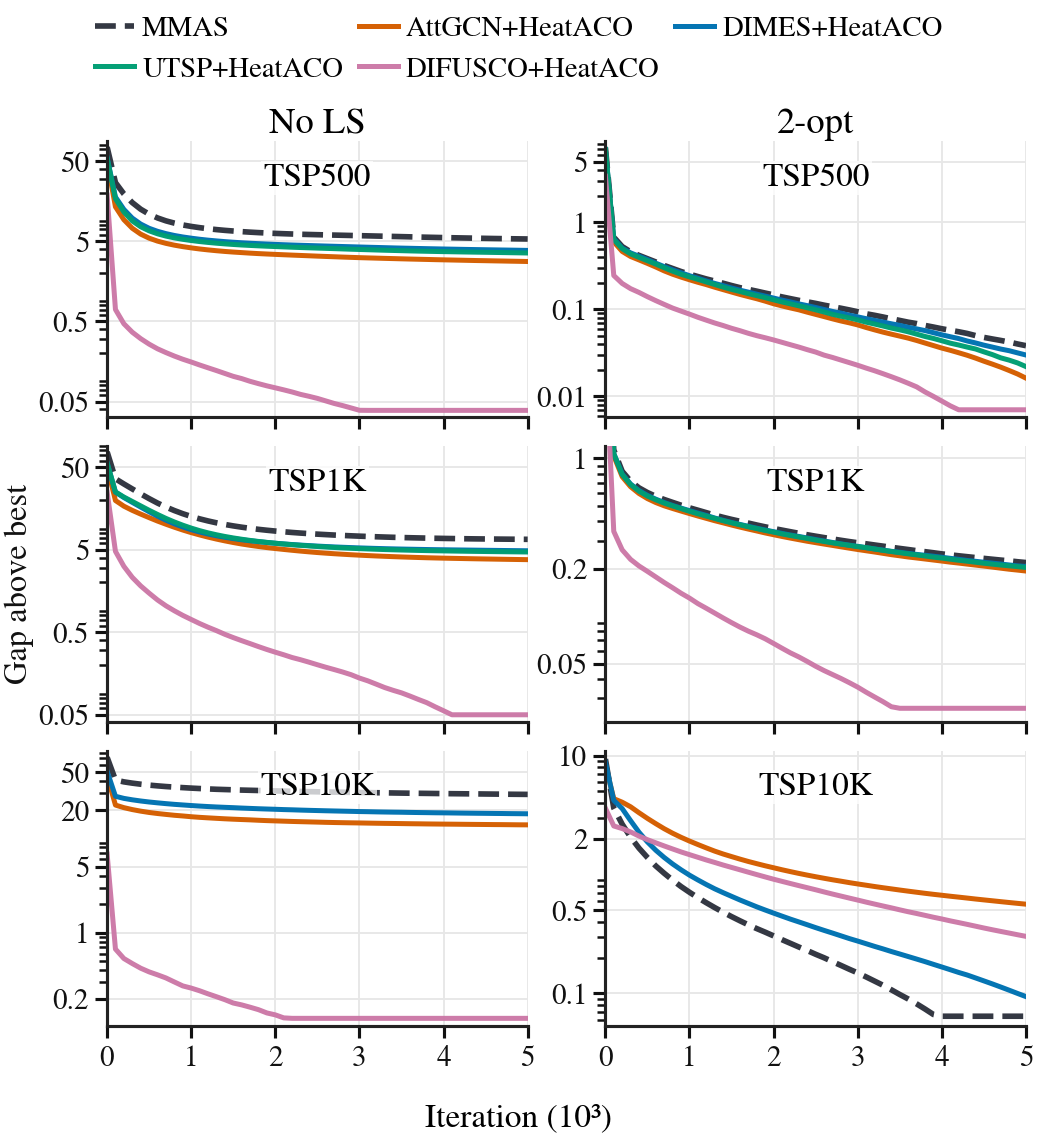}
  \caption{Mean best-so-far convergence on TSP500/TSP1K/TSP10K. Rows correspond to instance scale and columns to no local search (left) and 2-opt (right). The $y$-axis shows the positive excess gap above the best final mean gap within each panel on a logarithmic scale. Legend entries show predictor+\HeatACO.}
  \vspace{-2mm}
  \label{fig:convergence}
\end{figure}

Fig.~\ref{fig:convergence} shows the convergence behaviour. Without local search, heatmap-guided curves fall quickly during early iterations at all three scales, whereas standard \MMAS{} needs more iterations to discover comparable edge combinations. This supports the role of the evidence bias: it does not replace pheromone learning, but changes the initial sampling distribution so that pheromone feedback starts from a better part of the search space. With 2-opt, the curves are closer because local repair removes many construction errors. At TSP10K they also expose the interaction seen in Table~\ref{tab:main}: reliable heatmap guidance improves the unrefined tours, but the matched geometric baseline reaches a lower final gap once every constructed tour is repaired.

\subsection{Out-of-Distribution Generalisation}
\label{subsec:ood_generalisation}

Under spatial distribution shift, the heatmap continues to help when it stays reliable and stops helping when it degrades, and the clustered and Gaussian benchmarks in Table~\ref{tab:ood_clustered_gaussian} separate these cases. On Clustered TSP500 the matched \MMAS{}+2-opt baseline is already strong at $0.07\%$, and \HeatACO{}+2-opt does not improve on it, but it stays in the same range at $0.11\%$ to $0.23\%$ and remains far ahead of MCTS on the same shifted heatmaps, where AttGCN reaches $0.11\%$ against $1.37\%$ and DIMES reaches $0.17\%$ against $6.14\%$. On Gaussian TSP500 the outcome depends on the predictor. The AttGCN and DIMES heatmaps still improve on \MMAS{}+2-opt, at $0.16\%$ and $0.21\%$ against $0.23\%$, whereas the UTSP and DIFUSCO heatmaps degrade, and UTSP is the worst at $1.98\%$. The decoder therefore transfers across spatial shift, but the magnitude and direction of its benefit track the reliability of the shifted heatmap, which Section~\ref{sec:further_analysis} quantifies.

\begin{table}[!t]
  \centering
  \caption{OOD Results on Clustered and Gaussian TSP500.}
  \label{tab:ood_clustered_gaussian}
  \TableFont
  \begin{tabular}{@{}c|cc|cc@{}}
    \hline
    \multirow{2}{*}{\textbf{Method}} & \multicolumn{2}{c|}{\textbf{Clustered TSP500}} & \multicolumn{2}{c}{\textbf{Gaussian TSP500}} \\
    & MCTS & \HeatACO{}+2-opt & MCTS & \HeatACO{}+2-opt \\
    \hline
    MMAS+2-opt & --- & \textbf{0.07{\footnotesize\textcolor{gray}{$\pm$0.08}}} & --- & 0.23{\footnotesize\textcolor{gray}{$\pm$0.08}} \\
    AttGCN  & 1.37 & 0.11{\footnotesize\textcolor{gray}{$\pm$0.10}}\SigWorse & 1.14 & \textbf{0.16{\footnotesize\textcolor{gray}{$\pm$0.05}}}\SigBetter \\
    DIMES   & 6.14 & 0.17{\footnotesize\textcolor{gray}{$\pm$0.16}}\SigWorse & 5.33 & \textbf{0.21{\footnotesize\textcolor{gray}{$\pm$0.07}}}\SigBetter \\
    UTSP    & 4.85 & 0.23{\footnotesize\textcolor{gray}{$\pm$0.20}}\SigWorse & 190.12 & 1.98{\footnotesize\textcolor{gray}{$\pm$0.29}}\SigWorse \\
    DIFUSCO & 3.33 & 0.15{\footnotesize\textcolor{gray}{$\pm$0.14}}\SigWorse & 7.14 & 0.48{\footnotesize\textcolor{gray}{$\pm$0.19}}\SigWorse \\
    \hline
  \end{tabular}

  \vspace{2pt}
  \begin{minipage}{\columnwidth}
  \TableNoteFont\emph{Note:} Cells report mean$\pm$std gap (\%); std is computed over instance-level mean gaps after averaging seeds. \SigBetter{} indicates significant improvement over the corresponding \MMAS{} baseline ($p<0.05$); \SigWorse{} indicates significant degradation; \SigTie{} indicates no significant difference.
  \end{minipage}
  \vspace{-4mm}
\end{table}

\begin{table}[!t]
  \centering
  \caption{Generalisation to TSPLIB Instances.}
  \label{tab:tsplib}
  \TableFont
  \setlength{\tabcolsep}{2.8pt}
  \begin{tabular}{@{}cc|c|c|c@{}}
    \hline
    \multirow{2}{*}{\textbf{Method}} & \multirow{2}{*}{\textbf{Type}} & \textbf{pcb442} & \textbf{pr1002} & \textbf{pr2392} \\
    & & Gap (\%) & Gap (\%) & Gap (\%) \\
    \hline
    \multirow{3}[1]{*}{MMAS} & H & 5.32{\footnotesize\textcolor{gray}{$\pm$0.87}} & 7.14{\footnotesize\textcolor{gray}{$\pm$1.82}} & 23.86{\footnotesize\textcolor{gray}{$\pm$6.59}} \\
          & H+2-opt & 0.30{\footnotesize\textcolor{gray}{$\pm$0.08}} & 0.62{\footnotesize\textcolor{gray}{$\pm$0.17}} & 0.78{\footnotesize\textcolor{gray}{$\pm$0.18}} \\
          & H+3-opt & 0 & 0.09{\footnotesize\textcolor{gray}{$\pm$0.10}} & 0.23{\footnotesize\textcolor{gray}{$\pm$0.10}} \\
    MCTS  & MCTS  & 0.31 & 2.60 & 1.77 \\
    \hline
    AttGCN  & \multirow{4}[2]{*}{\rotatebox[origin=c]{90}{\shortstack[c]{NAR\\+Greedy}}} & 40.33 & 73.94 & 90.24 \\
    DIMES   &       & 70.80 & 116.01 & 180.83 \\
    UTSP    &       & 20.03 & 18.06 & 21.67 \\
    DIFUSCO &       & 27.40 & 21.34 & 18.89 \\
    \hline
    AttGCN  & \multirow{4}[2]{*}{\rotatebox[origin=c]{90}{\scriptsize+\HeatACO}} & 2.74{\footnotesize\textcolor{gray}{$\pm$0.51}}\SigBetter & 7.71{\footnotesize\textcolor{gray}{$\pm$2.59}}\SigTie & 17.47{\footnotesize\textcolor{gray}{$\pm$1.79}}\SigBetter \\
    DIMES   &       & 3.37{\footnotesize\textcolor{gray}{$\pm$0.63}}\SigBetter & 30.29{\footnotesize\textcolor{gray}{$\pm$2.44}}\SigWorse & 89.17{\footnotesize\textcolor{gray}{$\pm$1.18}}\SigWorse \\
    UTSP    &       & 4.12{\footnotesize\textcolor{gray}{$\pm$0.52}}\SigBetter & 13.83{\footnotesize\textcolor{gray}{$\pm$4.67}}\SigWorse & 33.01{\footnotesize\textcolor{gray}{$\pm$1.50}}\SigWorse \\
    DIFUSCO &       & 1.29{\footnotesize\textcolor{gray}{$\pm$0.19}}\SigBetter & 3.69{\footnotesize\textcolor{gray}{$\pm$0.27}}\SigBetter & 10.08{\footnotesize\textcolor{gray}{$\pm$0.98}}\SigBetter \\
    \hline
    AttGCN  & \multirow{4}[2]{*}{\rotatebox[origin=c]{90}{+MCTS}} & 0.24 & 2.43 & 2.77 \\
    DIMES   &       & 0.25 & 1.59 & 2.37 \\
    UTSP    &       & 0.56 & 1.94 & 1.86 \\
    DIFUSCO &       & 0.26 & 1.32 & 2.54 \\
    \hline
    AttGCN  & \multirow{4}[2]{*}{\rotatebox[origin=c]{90}{\scriptsize\shortstack[c]{+\HeatACO\\+2-opt}}} & 0.31{\footnotesize\textcolor{gray}{$\pm$0.07}}\SigTie & 0.73{\footnotesize\textcolor{gray}{$\pm$0.20}}\SigWorse & \textbf{0.75{\footnotesize\textcolor{gray}{$\pm$0.17}}}\SigTie \\
    DIMES   &       & \textbf{0.27{\footnotesize\textcolor{gray}{$\pm$0.09}}}\SigBetter & 2.32{\footnotesize\textcolor{gray}{$\pm$0.21}}\SigWorse & 4.08{\footnotesize\textcolor{gray}{$\pm$0.22}}\SigWorse \\
    UTSP    &       & 0.30{\footnotesize\textcolor{gray}{$\pm$0.02}}\SigTie & 0.77{\footnotesize\textcolor{gray}{$\pm$0.16}}\SigWorse & 1.43{\footnotesize\textcolor{gray}{$\pm$0.27}}\SigWorse \\
    DIFUSCO &       & 0.29{\footnotesize\textcolor{gray}{$\pm$0.02}}\SigTie & \textbf{0.53{\footnotesize\textcolor{gray}{$\pm$0.18}}}\SigBetter & 0.77{\footnotesize\textcolor{gray}{$\pm$0.16}}\SigTie \\
    \hline
    AttGCN  & \multirow{4}[2]{*}{\rotatebox[origin=c]{90}{\scriptsize\shortstack[c]{+\HeatACO\\+3-opt}}} & \textbf{0}\SigTie & 0.07{\footnotesize\textcolor{gray}{$\pm$0.08}}\SigTie & \textbf{0.17{\footnotesize\textcolor{gray}{$\pm$0.08}}}\SigBetter \\
    DIMES   &       & \textbf{0}\SigTie & 0.52{\footnotesize\textcolor{gray}{$\pm$0.11}}\SigWorse & 1.55{\footnotesize\textcolor{gray}{$\pm$0.16}}\SigWorse \\
    UTSP    &       & 0.0016{\footnotesize\textcolor{gray}{$\pm$0.0065}}\SigTie & 0.21{\footnotesize\textcolor{gray}{$\pm$0.10}}\SigWorse & 0.30{\footnotesize\textcolor{gray}{$\pm$0.11}}\SigWorse \\
    DIFUSCO &       & 0.01{\footnotesize\textcolor{gray}{$\pm$0.05}}\SigTie & \textbf{0}\SigBetter & 0.23{\footnotesize\textcolor{gray}{$\pm$0.10}}\SigTie \\
    \hline
  \end{tabular}

  \vspace{2pt}
  \begin{minipage}{\columnwidth}
  \TableNoteFont\emph{Note:} Gap measured against the TSPLIB reference. H denotes construction without local search. \MMAS{} and \HeatACO{} entries report mean$\pm$std over 30 seeds, with exact zero shown as 0; Greedy and MCTS entries report published mean gaps. \SigBetter{} indicates significant improvement, \SigWorse{} significant degradation, and \SigTie{} no significant difference over matched \MMAS{} across the 30 paired seeds at $p<0.05$.
  \end{minipage}
  \vspace{-4mm}
\end{table}

The TSPLIB instances in Table~\ref{tab:tsplib} provide the strongest test of structural transfer, because they have application-derived geometry rather than synthetic uniform sampling, and use the standard \texttt{EUC\_2D} rounding rule. Greedy heatmap decoding is weak here, with gaps from $18.06\%$ to $180.83\%$, while \HeatACO{} recovers usable tours from the same fixed heatmaps. On pcb442 the construction-only gaps fall to $1.29\%$--$4.12\%$ and \HeatACO{}+2-opt reaches $0.27\%$--$0.31\%$; on the larger instances the best heatmap-guided variants remain strong, with DIFUSCO+\HeatACO{}+2-opt at $0.53\%$ on pr1002 and AttGCN+\HeatACO{}+2-opt at $0.75\%$ on pr2392. The best 3-opt entries reach $0\%$ on pcb442, $0\%$ on pr1002, and $0.17\%$ on pr2392. Table~\ref{tab:tsplib} also exposes the most pronounced failure cases we observe: DIMES+\HeatACO{} stays far worse than \MMAS{} on pr1002 and pr2392, and UTSP degrades on the larger instances. These failures are predictor-specific, indicating that the cause lies in the heatmap, not the decoder; Section~\ref{subsec:tsplib_failure} traces them to candidate-density and confidence-separation failures of the heatmap itself.

\subsection{Directed ATSP and Heatmap Transfer}
\label{subsec:heatmap_sensitivity}

\begin{table}[!t]
  \centering
  \caption{ATSP Results with MatDIFFNet Heatmaps.}
  \label{tab:atsp}
  \TableFont
  \begin{tabular}{@{}cc|cc@{}}
    \hline
    \multirow{2}{*}{\textbf{Method}} & \multirow{2}{*}{\textbf{Type}} & \textbf{ATSP50} & \textbf{ATSP100} \\
    & & Gap (\%) & Gap (\%) \\
    \hline
    Gurobi~\cite{maML4COBench101BenchmarkMachine2025} & Exact          & --- & --- \\
    \multirow{2}[0]{*}{MMAS} & Heuristic      & 2.65{\footnotesize\textcolor{gray}{$\pm$0.02}}  & 6.15{\footnotesize\textcolor{gray}{$\pm$0.02}} \\
                              & Heuristic+2-opt & 0.62{\footnotesize\textcolor{gray}{$\pm$0.01}}  & 3.76{\footnotesize\textcolor{gray}{$\pm$0.02}} \\
    \hline
    \multirow{2}[0]{*}{MatDIFFNet} & NAR      & 33.2 & 24.1 \\
                                   & NAR+2-opt & 10.6 & 9.6  \\
    \hline
    \multirow{2}[0]{*}{MatDIFFNet} & \textbf{+\HeatACO}      & 0.81{\footnotesize\textcolor{gray}{$\pm$0.006}}\SigBetter & 1.07{\footnotesize\textcolor{gray}{$\pm$0.006}}\SigBetter \\
                                   & \textbf{+\HeatACO{}+2-opt} & \textbf{0.23{\footnotesize\textcolor{gray}{$\pm$0.005}}}\SigBetter & \textbf{0.57{\footnotesize\textcolor{gray}{$\pm$0.005}}}\SigBetter \\
    \hline
  \end{tabular}

  \vspace{2pt}
  \begin{minipage}{\columnwidth}
  \TableNoteFont\emph{Note:} Gaps computed against the Gurobi optimal tours supplied by ML4CO-Bench-101~\cite{maML4COBench101BenchmarkMachine2025}; dashes in the Gurobi row denote the defining reference gap. \SigBetter{} indicates significant improvement, and \SigTie{} no significant difference, over matched \MMAS{} ($p<0.05$).
  \end{minipage}
  \vspace{-4mm}
\end{table}

The clearest evidence that the heatmap helps where the available local search is weak comes from the asymmetric TSP. In our \ATSP{} experiments, local search is limited to a directed 2-opt operator, which is a much weaker repair than the symmetric 2-opt and 3-opt used above, so more of the burden of finding tour structure falls on construction. Table~\ref{tab:atsp} reports results with directed heatmaps from MatDIFFNet, the diffusion-based matrix-encoded TSP solver proposed in UniCO~\cite{pan_unico_2024} to extend heatmap prediction to asymmetric and non-Euclidean instances. Construction-only \HeatACO{} reaches $0.81\%$ on ATSP50 and $1.07\%$ on ATSP100, far below both the native MatDIFFNet decoder, at $33.2\%$ and $24.1\%$, and standard \MMAS, at $2.65\%$ and $6.15\%$. With the directed 2-opt enabled, \HeatACO{}+2-opt reaches $0.23\%$ and $0.57\%$, and it stays significantly ahead of \MMAS{}+2-opt, at $0.62\%$ and $3.76\%$. Unlike the symmetric case, where strong local search subsumes much of the heatmap's contribution, here the heatmap remains essential even with local search, because the directed 2-opt cannot recover the structure on its own. Adapting the decoder to \ATSP{} requires only changing the degree target, the evaporation rate, and the local-search operator, which shows that the evidence bias is not tied to symmetric edge selection.

GenSCO builds degree-awareness into the predictor through a constraint-sensitive output stage, so its heatmaps already concentrate on near-feasible tour structure. \HeatACO{} therefore acts here as a refinement step on top of an existing solver rather than as its replacement (Table~\ref{tab:gensco_alt}). On in-distribution TSP1K the refinement matches the native pipeline without improving on it, with \HeatACO{}+2-opt at $0.06\%$ against $0.04\%$ for GenSCO+2-opt. The strong in-distribution behaviour does not transfer, however, and this is where \HeatACO{} matters most. On Clustered TSP500, \HeatACO{}+2-opt lowers the original GenSCO gap from $5.62\%$ to $0.06\%$, ahead of the native GenSCO+2-opt at $0.48\%$. On Gaussian TSP500 the native GenSCO output collapses to $654.23\%$ and its own 2-opt leaves it at $558.50\%$, a near-total failure, whereas \HeatACO{} brings the same heatmap to $31.21\%$ without local search and to $3.00\%$ with 2-opt. \HeatACO{} recovers most of this lost quality: pheromone feedback repairs an informative but globally inconsistent heatmap, which the native one-shot decoder cannot.

\begin{table}[!t]
  \centering
  \caption{GenSCO Heatmap Transfer Results.}
  \label{tab:gensco_alt}
  \TableFont
  \setlength{\tabcolsep}{4.5pt}
  \begin{tabular}{@{}c|cccc@{}}
    \hline
    \textbf{Setting} & Original & +2-opt & +\HeatACO & +\HeatACO{}+2-opt \\
    \hline
    TSP1K (in-dist.) & 0.19 & 0.04 & 0.16{\footnotesize\textcolor{gray}{$\pm$0.14}} & 0.06{\footnotesize\textcolor{gray}{$\pm$0.04}} \\
    TSP500 Clustered & 5.62 & 0.48 & 1.37{\footnotesize\textcolor{gray}{$\pm$1.21}} & \textbf{0.06{\footnotesize\textcolor{gray}{$\pm$0.09}}} \\
    TSP500 Gaussian  & 654.23 & 558.50 & 31.21{\footnotesize\textcolor{gray}{$\pm$4.96}} & \textbf{3.00{\footnotesize\textcolor{gray}{$\pm$0.33}}} \\
    \hline
  \end{tabular}

  \vspace{2pt}
  \begin{minipage}{\columnwidth}
  \TableNoteFont\emph{Note:} GenSCO $C{=}160$ configuration. Cells report mean$\pm$std gap (\%); std is computed over instance-level mean gaps after averaging seeds. Bold values mark the best result for each setting.
  \end{minipage}
  \vspace{-2mm}
\end{table}

\begin{table}[!t]
  \centering
  \caption{SoftDist Heatmap Transfer Results.}
  \label{tab:softdist_transfer}
  \TableFont
  \setlength{\tabcolsep}{4pt}
  \begin{tabular}{@{}c|cc|cc|cc@{}}
    \hline
    \multirow{2}{*}{\textbf{Settings}} & \multicolumn{2}{c|}{+MCTS} & \multicolumn{2}{c|}{+\HeatACO} & \multicolumn{2}{c}{+\HeatACO{}+2-opt} \\
    & Gap & Time & Gap & Time & Gap & Time \\
    \hline
    TSP500  & 0.43 & 50\,s & 4.19{\footnotesize\textcolor{gray}{$\pm$0.65}} & 1.64\,s & 0.164{\footnotesize\textcolor{gray}{$\pm$0.122}} & 1.95\,s \\
    TSP1K   & 3.13 & 100\,s & 5.51{\footnotesize\textcolor{gray}{$\pm$0.60}} & 3.98\,s & 0.425{\footnotesize\textcolor{gray}{$\pm$0.137}} & 4.30\,s \\
    TSP10K  & 3.13 & 16.67\,m & 24.72{\footnotesize\textcolor{gray}{$\pm$0.83}} & 5.90\,m & 0.540{\footnotesize\textcolor{gray}{$\pm$0.096}} & 60.92\,s \\
    \hline
  \end{tabular}

  \vspace{2pt}
  \begin{minipage}{\columnwidth}
  \TableNoteFont\emph{Note:} SoftDist is a non-learned heatmap formed by a softmax over the distance matrix, so it encodes only proximity. Gap cells report mean$\pm$std where repeated-run statistics are available; MCTS values are published mean gap/time.
  \vspace{-4mm}
  \end{minipage}
\end{table}

As a non-learned control, Table~\ref{tab:softdist_transfer} uses SoftDist, which produces a heatmap by applying a softmax to the (negated) distance matrix, so the resulting confidence is a monotone function of pairwise distance alone. It is therefore not a learned signal at all and carries no information beyond the geometry that \HeatACO{} already sees through its distance heuristic $\eta_{ij}=1/d_{ij}$. The decoder remains numerically stable on this degenerate input, and with 2-opt it remains strong at every scale, reaching $0.16\%$, $0.43\%$, and $0.54\%$ on TSP500, TSP1K, and TSP10K. The TSP10K figure is the most informative, because there SoftDist with 2-opt is on par with the learned heatmaps and far ahead of SoftDist with MCTS at $3.13\%$. This is a property of the setting, not of the decoder: at large scale with strong local search, a purely geometric prior is already a strong and consistent signal, the same setting where the learned heatmaps lose their advantage. The control therefore supports the central observation: a learned heatmap is worth using only when it supplies reliable structure beyond pairwise distance, and elsewhere a geometric prior plus local search is hard to outperform.

\subsection{Ablation and Sensitivity}
\label{subsec:parameter_sensitivity}

We evaluate the design of the evidence bias through two ablation studies. The first removes its two scaling components in turn, and the second varies the single free parameter of the cap rule.

\begin{table}[!t]
  \centering
  \caption{TSP1K Evidence-Bias Component Ablation.}
  \label{tab:ablation_components}
  \TableFont
  \setlength{\tabcolsep}{1.5pt}
  \begin{tabular}{@{}cc|ccc@{}}
    \hline
    \textbf{Heatmap} & \textbf{Variant} & \textbf{No LS} & \textbf{2-opt} & \textbf{3-opt} \\
    \hline
    \multirow{3}{*}{AttGCN}
      & Main & \textbf{4.85{\footnotesize\textcolor{gray}{$\pm$0.31}}} & \textbf{0.39{\footnotesize\textcolor{gray}{$\pm$0.06}}} & \textbf{0.053{\footnotesize\textcolor{gray}{$\pm$0.021}}} \\
      & Fixed $\kappa{=}1$ & 6.67{\footnotesize\textcolor{gray}{$\pm$0.22}}\SigWorse & 0.42{\footnotesize\textcolor{gray}{$\pm$0.05}}\SigWorse & 0.059{\footnotesize\textcolor{gray}{$\pm$0.021}}\SigWorse \\
      & No cap & 5.42{\footnotesize\textcolor{gray}{$\pm$0.23}}\SigWorse & 0.40{\footnotesize\textcolor{gray}{$\pm$0.05}}\SigWorse & 0.058{\footnotesize\textcolor{gray}{$\pm$0.020}}\SigWorse \\
    \hline
    \multirow{3}{*}{DIMES}
      & Main & \textbf{5.91{\footnotesize\textcolor{gray}{$\pm$0.21}}} & \textbf{0.40{\footnotesize\textcolor{gray}{$\pm$0.05}}} & \textbf{0.056{\footnotesize\textcolor{gray}{$\pm$0.020}}} \\
      & Fixed $\kappa{=}1$ & 6.79{\footnotesize\textcolor{gray}{$\pm$0.34}}\SigWorse & 0.41{\footnotesize\textcolor{gray}{$\pm$0.05}}\SigWorse & 0.059{\footnotesize\textcolor{gray}{$\pm$0.019}}\SigWorse \\
      & No cap & 6.26{\footnotesize\textcolor{gray}{$\pm$0.20}}\SigWorse & 0.40{\footnotesize\textcolor{gray}{$\pm$0.05}}\SigTie & 0.057{\footnotesize\textcolor{gray}{$\pm$0.021}}\SigTie \\
    \hline
    \multirow{3}{*}{UTSP}
      & Main & \textbf{5.75{\footnotesize\textcolor{gray}{$\pm$0.25}}} & \textbf{0.40{\footnotesize\textcolor{gray}{$\pm$0.05}}} & \textbf{0.055{\footnotesize\textcolor{gray}{$\pm$0.021}}} \\
      & Fixed $\kappa{=}1$ & 7.72{\footnotesize\textcolor{gray}{$\pm$0.21}}\SigWorse & 0.42{\footnotesize\textcolor{gray}{$\pm$0.05}}\SigWorse & 0.060{\footnotesize\textcolor{gray}{$\pm$0.021}}\SigWorse \\
      & No cap & 6.06{\footnotesize\textcolor{gray}{$\pm$0.21}}\SigWorse & 0.41{\footnotesize\textcolor{gray}{$\pm$0.05}}\SigTie & 0.057{\footnotesize\textcolor{gray}{$\pm$0.022}}\SigWorse \\
    \hline
    \multirow{3}{*}{DIFUSCO}
      & Main & \textbf{1.07{\footnotesize\textcolor{gray}{$\pm$0.38}}} & \textbf{0.20{\footnotesize\textcolor{gray}{$\pm$0.07}}} & \textbf{0.040{\footnotesize\textcolor{gray}{$\pm$0.026}}} \\
      & Fixed $\kappa{=}1$ & 1.11{\footnotesize\textcolor{gray}{$\pm$0.35}}\SigWorse & 0.21{\footnotesize\textcolor{gray}{$\pm$0.06}}\SigWorse & 0.040{\footnotesize\textcolor{gray}{$\pm$0.023}}\SigTie \\
      & No cap & 4.26{\footnotesize\textcolor{gray}{$\pm$0.18}}\SigWorse & 0.37{\footnotesize\textcolor{gray}{$\pm$0.05}}\SigWorse & 0.058{\footnotesize\textcolor{gray}{$\pm$0.021}}\SigWorse \\
    \hline
  \end{tabular}

  \vspace{2pt}
  \begin{minipage}{\columnwidth}
  \TableNoteFont\emph{Note:} Bold values are best within each block. Std is computed over instance-level mean gaps after averaging seeds. \SigWorse{} marks a significant degradation from Main, and \SigTie{} no significant difference, at $p<0.05$ (paired Wilcoxon on instance-level mean deltas).
  \end{minipage}
  \vspace{-4mm}
\end{table}

Table~\ref{tab:ablation_components} isolates the two components of the scaling rule on TSP1K. Fixing $\kappa=1$ removes the instance-level adaptation, and ``No cap'' removes the pheromone-range constraint. The full method is best, or statistically tied for best, in almost every heatmap and local-search block, and the effect is clearest in construction, where no local search can mask a scaling error. Removing the cap is most damaging on the strongest heatmap, where it degrades DIFUSCO from $1.07\%$ to $4.26\%$, because a sharply separated heatmap then produces an unbounded factor that overpowers the pheromone feedback. Fixing the scale is most damaging on the weaker heatmaps, where it degrades AttGCN from $4.85\%$ to $6.67\%$ and UTSP from $5.75\%$ to $7.72\%$, because a uniform scale is too aggressive for evidence that is faint and poorly separated. The automatic cap therefore handles the strong-heatmap risk and the automatic scale handles the weak-heatmap risk, so both are needed for the rule to work across predictors. Under 2-opt and 3-opt the differences shrink but stay significant in most blocks, for the same reason that the main gains shrink under stronger local search.

\begin{figure}[!t]
  \centering
  \includegraphics[width=\columnwidth]{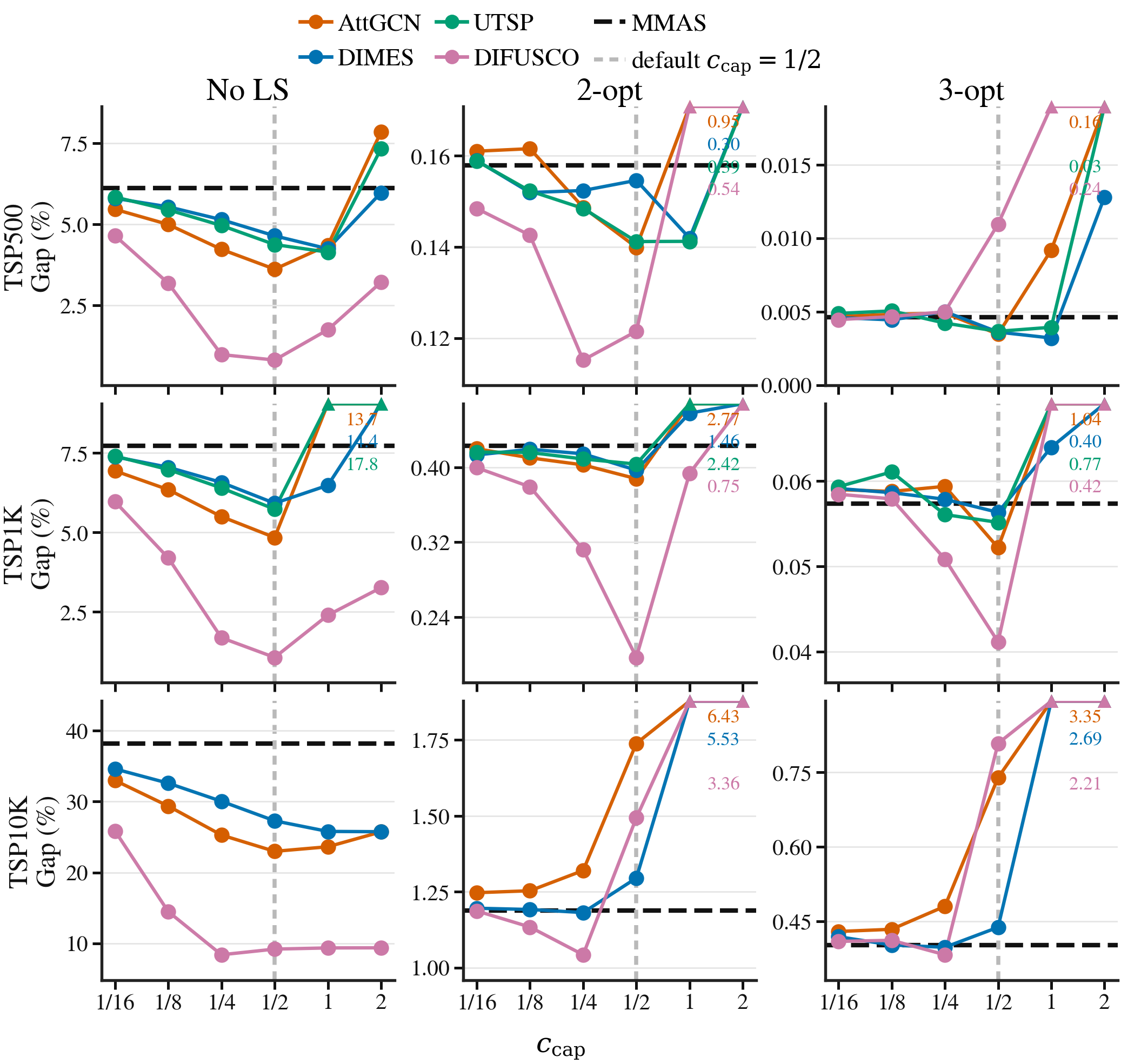}
  \caption{Sensitivity to the evidence cap factor $c_{\mathrm{cap}}$ on regular TSP benchmarks. Each panel reports the raw mean gap for the indicated heatmap-guided decoder and local search setting. The dark horizontal dashed line marks the standard \MMAS{} baseline, and the light-grey vertical dashed line marks the default $c_{\mathrm{cap}}=0.5$. Top-edge triangles indicate values clipped by the adaptive display range, with the true gap annotated nearby.}
  \vspace{-4mm}
  \label{fig:cap_sensitivity}
\end{figure}

The cap factor $c_{\mathrm{cap}}$ is the only free parameter introduced by the scaling rule, and Fig.~\ref{fig:cap_sensitivity} sweeps it over $\{0.0625,0.125,0.25,0.5,1,2\}$, from a weak bias to one as large as the pheromone range. The stable region is broad, and values from $0.125$ to $0.5$ behave similarly across most TSP500 and TSP1K settings. Very small caps underuse an informative heatmap, and very large caps can hurt construction because the bias dominates before the pheromone feedback has accumulated instance-specific evidence. Under 3-opt the curves are flatter, since local search compensates for part of the construction error. The default $c_{\mathrm{cap}}=0.5$ corresponds to half the log pheromone range.

\section{Heatmap Reliability and Decoding Limits}
\label{sec:further_analysis}

We now inspect the thresholded heatmap graph before per-node candidate truncation. The goal is to explain why the gains are stable during construction from TSP500 through TSP10K, become less complementary to strong local search at TSP10K, and remain predictor-dependent under distribution shift. 

For \HeatACO{} to improve on standard \MMAS, the heatmap must satisfy several conditions. The thresholded graph must contain most edges of a good tour, or the decoder cannot sample them. The number of surviving edges must stay moderate, or construction faces too many false alternatives. The useful edges must stand out above the row-level degree baseline, or the evidence factors in~\eqref{eq:evidence} approach one and the decoder falls back to distance and pheromone. Known-tour coverage is available only as a post-hoc analysis. Candidate density and row-level score separation can be evaluated from the heatmap alone, although whether the separated edges are useful remains a post-hoc question.

\subsection{Candidate Sparsity and Known-Tour Recall}
\label{subsec:candidate_diagnostics}

Let $E_{\theta_h}(H)=\{(i,j): H_{ij}>\theta_h\}$ be the directed heatmap entries surviving the global threshold $\theta_h=10^{-4}$ \cite{fu_generalize_2021}, before per-node truncation, and let $\bar E_{\theta_h}(H)$ be their undirected projection. We use three diagnostics: undirected-equivalent candidate density $\mathrm{Edges}/n=|E_{\theta_h}(H)|/(2n)$, known-tour coverage $\mathrm{Cov}(H)=|\bar E_{\theta_h}(H)\cap E^\star|/|E^\star|$, and the missing-edge percentage $\mathrm{Miss}(H)=100(1-\mathrm{Cov}(H))$, where $E^\star$ is the edge set of the known tour: the Concorde optimum for TSP500/TSP1K, the benchmark-provided known optimum for TSP10K, and the TSPLIB reference tour for the TSPLIB instances. Candidate density approximates the branching burden imposed on the decoder, coverage measures whether good edges are available at all, and missing edges identify cases where even a perfect decoder restricted to the thresholded graph would be limited. We report these diagnostics wherever both a heatmap and the known tour are available; UTSP has no TSP10K heatmap in this benchmark release and is therefore left blank.

\begin{table}[!t]
  \centering
  \caption{Heatmap-reliability diagnostics and \HeatACO{} outcome vs.\ matched \MMAS.}
  \label{tab:diag_merged}
  \TableFont
  \setlength{\tabcolsep}{5pt}
  \begin{tabular}{@{}cc ccc cc@{}}
    \toprule
    \textbf{Heatmap} & \textbf{Set} & \textbf{Edges/$n$} & \textbf{Miss\%} & \textbf{P(e$>$1)} & \textbf{no-LS} & \textbf{+2-opt} \\
    \midrule
    \multicolumn{7}{@{}l}{\textit{uniform TSP}} \\
\multirow{3}{*}{AttGCN} & TSP500 & 4.78 & 0.84 & \cellcolor{tabGreen}0.75 & \cellcolor{tabGreen}$\uparrow$ & \cellcolor{tabGreen}$\uparrow$ \\
 & TSP1K & 4.86 & 0.89 & \cellcolor{tabGreen}0.74 & \cellcolor{tabGreen}$\uparrow$ & \cellcolor{tabGreen}$\uparrow$ \\
 & TSP10K & 5.75 & 1.3 & \cellcolor{tabGreen}0.73 & \cellcolor{tabGreen}$\uparrow$ & \cellcolor{tabRed}$\downarrow$ \\
\addlinespace[1.5pt]
\multirow{3}{*}{DIMES} & TSP500 & 3.67 & \cellcolor{tabOrange}3.40 & \cellcolor{tabGreen}0.71 & \cellcolor{tabGreen}$\uparrow$ & \cellcolor{tabGreen}$\uparrow$ \\
 & TSP1K & 3.82 & \cellcolor{tabOrange}3.06 & \cellcolor{tabGreen}0.71 & \cellcolor{tabGreen}$\uparrow$ & \cellcolor{tabGreen}$\uparrow$ \\
 & TSP10K & 2.50 & 2.2 & \cellcolor{tabGreen}0.64 & \cellcolor{tabGreen}$\uparrow$ & \cellcolor{tabRed}$\downarrow$ \\
\addlinespace[1.5pt]
\multirow{3}{*}{UTSP} & TSP500 & \cellcolor{tabOrange}11.39 & 0.00 & \cellcolor{tabGreen}0.65 & \cellcolor{tabGreen}$\uparrow$ & \cellcolor{tabGreen}$\uparrow$ \\
 & TSP1K & \cellcolor{tabOrange}11.34 & 0.00 & \cellcolor{tabYellow}0.61 & \cellcolor{tabGreen}$\uparrow$ & \cellcolor{tabGreen}$\uparrow$ \\
 & TSP10K & -- & -- & -- & -- & -- \\
\addlinespace[1.5pt]
\multirow{3}{*}{DIFUSCO} & TSP500 & 4.51 & 0.62 & \cellcolor{tabGreen}0.86 & \cellcolor{tabGreen}$\uparrow$ & \cellcolor{tabGreen}$\uparrow$ \\
 & TSP1K & 4.51 & 0.62 & \cellcolor{tabGreen}0.86 & \cellcolor{tabGreen}$\uparrow$ & \cellcolor{tabGreen}$\uparrow$ \\
 & TSP10K & 5.90 & 0.2 & \cellcolor{tabGreen}0.81 & \cellcolor{tabGreen}$\uparrow$ & \cellcolor{tabRed}$\downarrow$ \\
    \midrule
    \multicolumn{7}{@{}l}{\textit{TSPLIB}} \\
\multirow{3}{*}{AttGCN} & pcb442 & 4.65 & 0.68 & \cellcolor{tabGreen}0.74 & \cellcolor{tabGreen}$\uparrow$ & \cellcolor{tabYellow}$=$ \\
 & pr1002 & 4.72 & \cellcolor{tabOrange}2.40 & \cellcolor{tabGreen}0.71 & \cellcolor{tabYellow}$=$ & \cellcolor{tabRed}$\downarrow$ \\
 & pr2392 & 4.77 & 1.71 & \cellcolor{tabGreen}0.74 & \cellcolor{tabGreen}$\uparrow$ & \cellcolor{tabYellow}$=$ \\
\addlinespace[1.5pt]
\multirow{3}{*}{DIMES} & pcb442 & 3.75 & \cellcolor{tabOrange}2.26 & \cellcolor{tabGreen}0.72 & \cellcolor{tabGreen}$\uparrow$ & \cellcolor{tabGreen}$\uparrow$ \\
 & pr1002 & \cellcolor{tabRed}116.34 & \cellcolor{tabRed}7.58 & \cellcolor{tabGreen}0.70 & \cellcolor{tabRed}$\downarrow$ & \cellcolor{tabRed}$\downarrow$ \\
 & pr2392 & \cellcolor{tabRed}493.26 & \cellcolor{tabRed}16.30 & \cellcolor{tabGreen}0.72 & \cellcolor{tabRed}$\downarrow$ & \cellcolor{tabRed}$\downarrow$ \\
\addlinespace[1.5pt]
\multirow{3}{*}{UTSP} & pcb442 & \cellcolor{tabRed}40.68 & 0.00 & \cellcolor{tabYellow}0.63 & \cellcolor{tabGreen}$\uparrow$ & \cellcolor{tabYellow}$=$ \\
 & pr1002 & \cellcolor{tabRed}68.99 & 0.00 & \cellcolor{tabRed}0.42 & \cellcolor{tabRed}$\downarrow$ & \cellcolor{tabRed}$\downarrow$ \\
 & pr2392 & \cellcolor{tabRed}112.62 & 0.00 & \cellcolor{tabRed}0.38 & \cellcolor{tabRed}$\downarrow$ & \cellcolor{tabRed}$\downarrow$ \\
\addlinespace[1.5pt]
\multirow{3}{*}{DIFUSCO} & pcb442 & 3.83 & 1.58 & \cellcolor{tabGreen}0.72 & \cellcolor{tabGreen}$\uparrow$ & \cellcolor{tabYellow}$=$ \\
 & pr1002 & \cellcolor{tabOrange}9.56 & 0.40 & \cellcolor{tabGreen}0.73 & \cellcolor{tabGreen}$\uparrow$ & \cellcolor{tabGreen}$\uparrow$ \\
 & pr2392 & \cellcolor{tabRed}23.43 & 0.00 & \cellcolor{tabGreen}0.75 & \cellcolor{tabGreen}$\uparrow$ & \cellcolor{tabYellow}$=$ \\
    \bottomrule
  \end{tabular}

  \vspace{2pt}
  \begin{minipage}{\columnwidth}
  \TableNoteFont\raggedright\emph{Note:} Edges/$n$ and Miss\% diagnose the heatmap; P(e$>$1) is the fraction of edges with additional evidence that belong to the known tour, a post-hoc control on the selection rule. $\uparrow/=/\downarrow$ is \HeatACO{} significantly better\,/\,tied\,/\,worse than matched \MMAS{} (paired Wilcoxon, $p<0.05$) at no-LS and $+$2-opt. Shading: \colorbox{tabRed}{red}\,/\,\colorbox{tabOrange}{orange} flag degraded heatmap diagnostics; \colorbox{tabGreen}{green} marks healthy selection or a \HeatACO{} win.
  \end{minipage}
  \vspace{-4mm}
\end{table}

At $\theta_h=10^{-4}$, the predictors reduce the dense $O(n^2)$ edge set to $O(n)$ candidates while keeping high known-tour recall (Table~\ref{tab:diag_merged}, uniform-TSP block). Across the available in-distribution scales, DIFUSCO yields $4.5$--$5.9$ edges per node with under $1\%$ of known-tour edges missing, AttGCN yields $4.8$--$5.8$ edges per node while missing only $0.8\%$--$1.3\%$, and DIMES is sparser at $2.5$--$3.8$ edges per node while missing $2.2\%$--$3.4\%$. UTSP is denser at about $11.3$ edges per node and nearly lossless on TSP500 and TSP1K. The table also reports the fraction $P(e\!>\!1)$ of edges with additional evidence that belong to the known tour. This reference-dependent quantity is not used by the algorithm and serves only as a post-hoc control. On uniform TSP it stays between $0.61$ and $0.86$, whereas an arbitrary thresholded candidate belongs to the known tour with probability only $0.09$ to $0.26$, so the rule concentrates the known-tour edge signal by a factor of $2.7$ to $7.4$. The heatmap graph is thus sparse, contains nearly all known-tour edges through TSP10K, and is filtered by a rule that reliably keeps the good ones.

High recall alone is not enough. A tour uses exactly two edges per node, yet even a sparse heatmap offers several plausible edges at each node, so the decoder must select a globally consistent subset under the degree and subtour constraints rather than merely confirm that good edges exist. The pheromone feedback performs this selection: \HeatACO{} samples many feasible tours, reinforces the edges that take part in short tours, and evaporates the rest. The diagnostics therefore link the gap improvements to a specific mechanism: the heatmap supplies a high-recall candidate region, and pheromone feedback resolves the conflicts within that region.

\subsection{Confidence-Band Concentration}
\label{subsec:confidence_band}

\begin{figure}[!t]
  \centering
  \subfloat[In-distribution TSP1K]{\includegraphics[width=\columnwidth]{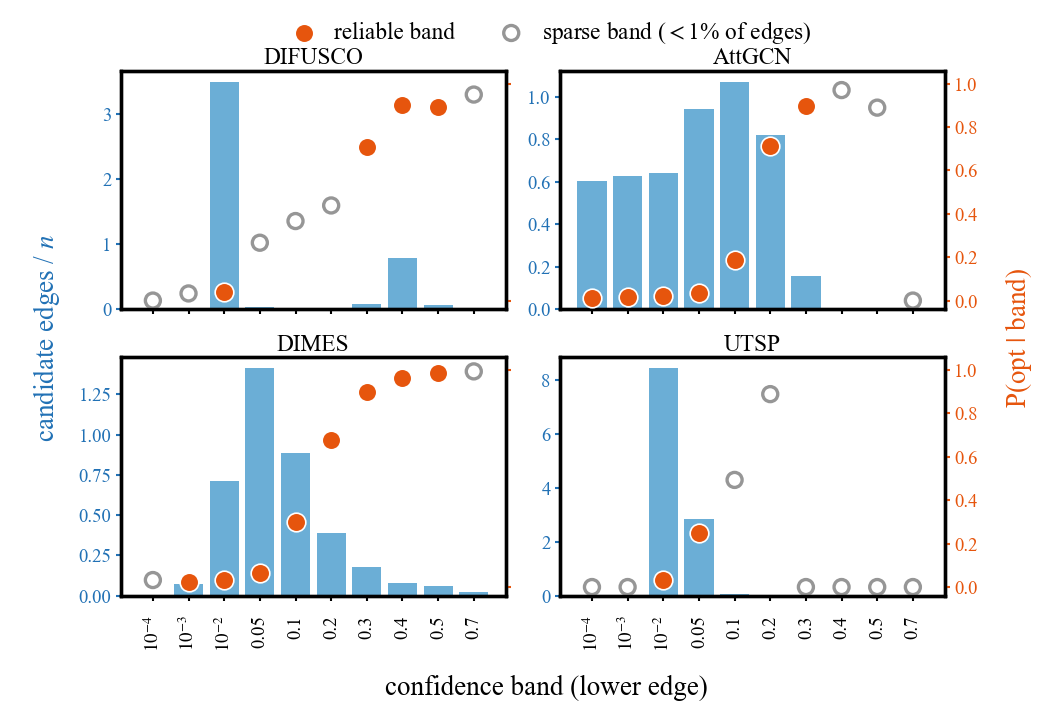}\label{fig:interval_contrib_indist}}\\
  \subfloat[TSPLIB pr2392]{\includegraphics[width=\columnwidth]{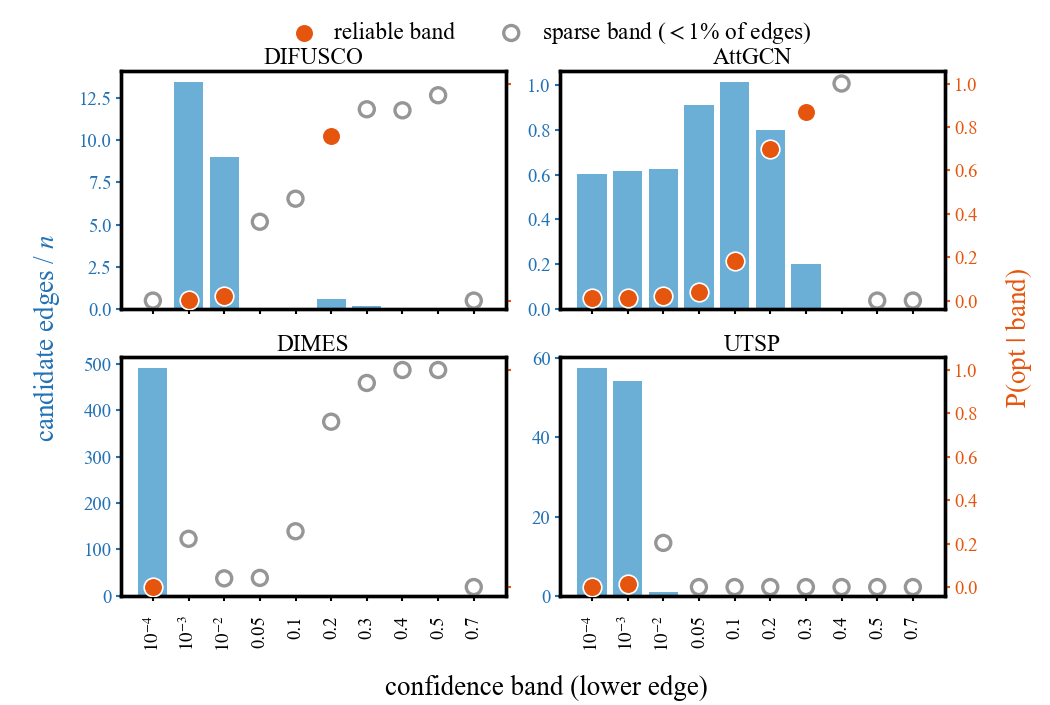}\label{fig:interval_contrib_tsplib}}
  \caption{Confidence-band separation on (a) in-distribution TSP1K and (b) the structurally shifted TSPLIB instance pr2392. Blue bars give the candidate edges per node in each confidence band (left axis), and orange markers give $P(\text{opt}\,|\,\text{band})$, the fraction of edges in that band that lie on the known tour (right axis, $0$--$1$). Filled markers are reliable bands holding at least $1\%$ of the predictor's candidate edges, and hollow markers are sparse bands.}
  \label{fig:interval_contrib}
  \vspace{-4mm}
\end{figure}

Fig.~\ref{fig:interval_contrib}(a) partitions the confidence values into bands and reports, for each band, the candidate-edge density and the conditional optimal-edge frequency $P(\text{opt}\,|\,\text{band})$. Higher confidence is not uniformly better. Known-tour edges concentrate in a narrow medium-to-high band of high discriminability, while the dense low-confidence bands are almost pure noise. For DIFUSCO the lowest populated band $[0.01,0.05)$ holds most of the candidate edges yet has $P(\text{opt}\,|\,\text{band})\approx0.04$, whereas the $[0.4,0.5)$ band holds about one edge per node at $P(\text{opt}\,|\,\text{band})\approx0.87$. DIMES has a smooth monotone relationship but spreads its optimal edges across several middle bands, and UTSP places about ten edges per node into low-confidence bands of near-zero discriminability.

This band structure motivates both the evidence formula and the cap. The degree baseline in~\eqref{eq:evidence} measures whether an edge stands out beyond the number of edges the node can use, so a node with many similar scores produces little evidence even at high raw confidence, because the heatmap exhibits low row-level discriminability there. The cap in~\eqref{eq:kappa} then bounds the strongest evidence relative to the pheromone range, which prevents a dense low-confidence band from dominating early construction by sheer volume. The method thus exploits the concentrated, well-separated band while keeping raw heatmap magnitude from overriding later pheromone correction.

The same band structure also clarifies the interaction with 2-opt and 3-opt. Heatmap guidance changes the starting tours and the pheromone trajectory, while local search itself always uses a separate geometric nearest-neighbour candidate list. When construction draws useful high-confidence edges, geometric repair starts from a better tour; however, once strong repair can recover the same structure independently, the heatmap's marginal benefit can vanish or reverse. TSP10K demonstrates this complementarity boundary directly: its heatmaps have healthy sparsity, recall, and selection precision, and they improve every unrefined instance, yet the matched geometric baseline attains the lower post-search gap.

\subsection{Failure Modes Under Distribution Shift}
\label{subsec:tsplib_failure}

Fig.~\ref{fig:interval_contrib}(b) and the TSPLIB block of Table~\ref{tab:diag_merged} show the effect of the geometry shift on the confidence mass. DIMES exhibits a sharp increase in low-confidence density: candidate density rises from $3.75$ edges per node on pcb442 to $116$ on pr1002 and $493$ on pr2392, with most of the extra mass in the lowest-confidence bins. The missing rate rises to $7.58\%$ and $16.30\%$, and known-tour edges become hard to separate from background candidates. The selection control $P(e\!>\!1)$ nonetheless stays around $0.72$, so the degree-aware rule still picks good edges, but they are diluted by a large volume of low-confidence candidates. This accounts for DIMES+\HeatACO{}+2-opt being competitive on pcb442, at $0.27\%$, but degrading on pr1002 and pr2392, at $2.32\%$ and $4.08\%$, where the failure comes from heatmap density, not from the selection rule.

UTSP fails in a related but distinct way. It keeps all known-tour edges, with zero missing in Table~\ref{tab:diag_merged}, but its thresholded graph is very dense at $40.68$ to $112.63$ edges per node, and its selection control also drops, with $P(e\!>\!1)$ falling to $0.42$ and $0.38$ on pr1002 and pr2392, so the decoder faces many competing low-signal candidates that the rule can no longer separate. AttGCN and DIFUSCO stay much sparser and keep $P(e\!>\!1)\ge0.71$. DIFUSCO still performs strongly on pr1002 with local search, at $0.53\%$ for 2-opt and $0\%$ for 3-opt, and AttGCN is strongest on pr2392, at $0.75\%$ for 2-opt and $0.17\%$ for 3-opt, so different predictors fail under different structural shifts.

\subsection{Boundaries of Heatmap-Guided Decoding}
\label{subsec:decoding_limits}

\HeatACO{} exhibits two empirical boundaries. The first is reliability: under severe distribution shift, a heatmap can become too dense or move its confidence mass away from known-tour edges, leaving little separable evidence. The evidence factors then approach one, although residual misleading candidates can still hurt on some TSPLIB combinations. The second is complementarity: at TSP10K the heatmaps remain sparse and high in coverage and improve every construction-only instance, but 2-opt and 3-opt recover much of the same geometric structure and the matched \MMAS{} variants finish lower. Heatmap quality therefore indicates whether guidance is usable, while local-search strength determines its additional value.

We do not claim that a neural heatmap should always outperform classical search. A fixed heatmap is useful when it carries well-separated edge evidence, and these diagnostics indicate when it does not.
\section{Conclusions and Future Work}
\label{sec:conclusion}

This paper addressed heatmap-to-tour decoding as a distinct stage of heatmap-based non-autoregressive solvers. We proposed \HeatACO, a scalable, predictor-agnostic decoder that adds a capped, degree-aware evidence factor to \MMAS{} and scales it from the pheromone dynamic range. \HeatACO{} retains fixed-size candidate sampling and parallel ant construction, while using the same evidence configuration across pretrained predictors without retraining or predictor-specific tuning. In every available in-distribution same-heatmap comparison, \HeatACO{} with 2-opt obtains lower gaps and shorter reported decoding times than the published MCTS baselines. Against matched \MMAS{} baselines on these benchmarks, it improves construction for every available predictor and scale and remains beneficial in most local-search settings on TSP500 and TSP1K. Directed \ATSP{} experiments show consistent gains, while distribution shifts reveal both successful transfer and predictor-specific failures. At TSP10K, its advantage over matched \MMAS{} is construction-specific, which is consistent with reduced complementarity under strong geometric repair. Together, these results support reusable fixed-heatmap decoding across large-scale benchmarks while showing that its value depends on heatmap reliability and complementarity with local search.

Future work should combine score-only heatmap reliability with local-search strength to select between \HeatACO{} and a full \MMAS{} fallback.

\bibliographystyle{IEEEtran}
\bibliography{heataco}

\end{document}